\documentclass[twoside,twocolumn,twocolumn]{IEEEtran}
\usepackage[T1]{fontenc}
\usepackage{color}
\usepackage{float}
\usepackage{amsmath}
\usepackage{amsthm}
\usepackage{amssymb}
\usepackage{graphicx}
\usepackage[normalem]{ulem}
\usepackage{comment}

\makeatletter

\floatstyle{ruled}
\newfloat{algorithm}{tbp}{loa}
\providecommand{\algorithmname}{Algorithm}
\floatname{algorithm}{\protect\algorithmname}

  \theoremstyle{plain}
  
  \theoremstyle{plain}
  \newtheorem{thm}{\protect\theoremname}
  \theoremstyle{plain}
  
  \theoremstyle{plain}
  
  \theoremstyle{plain}
  \newtheorem{rem}{\protect\remarkname}

\ifCLASSOPTIONcompsoc
\usepackage[caption=false,font=normalsize,labelfont=sf,textfont=sf]{subfig}
\else
\usepackage[caption=false,font=footnotesize]{subfig}
\fi

\usepackage{cite}
\usepackage{bm}
\usepackage{algorithmic}
\usepackage{algorithm}  
\usepackage{graphicx}
\usepackage{bbm}
\usepackage{mathrsfs}
\usepackage{enumitem}
\IEEEoverridecommandlockouts
\makeatother

\providecommand{\propositionname}{Proposition}
\providecommand{\corollaryname}{Corollary}
\providecommand{\theoremname}{Theorem}
\providecommand{\lemmaname}{Lemma}
\providecommand{\remarkname}{Remark}

\begin{document}


\title{Accelerating Wireless Distributed Learning via Hybrid Split and Federated Learning Optimization}
\author{Kun~Guo,~\IEEEmembership{Member,~IEEE}, Xuefei~Li, Xijun~Wang,~\IEEEmembership{Member,~IEEE}, Howard~H.~Yang,~\IEEEmembership{Member,~IEEE}, Wei~Feng,~\IEEEmembership{Senior Member,~IEEE}, and~Tony~Q. S.~Quek,~\IEEEmembership{Fellow,~IEEE}\thanks{This paper was accepted in part at the IEEE GLOBECOM 2024 \cite{GC_Xuefei}.} 
\thanks{K. Guo and X. Li are with the School of Communications and Electronics Engineering, East China Normal University, Shanghai 200241, China (e-mail: kguo@cee.ecnu.edu.cn, 51255904087@stu.ecnu.edu.cn)}
\thanks{X. Wang is with the School of Electronics and Information Technology, Sun Yat-sen University, Guangzhou 510006, China (e-mail: wangxijun@mail.sysu.edu.cn)}
\thanks{H. H. Yang is with the Zhejiang University/University of Illinois at Urbana-Champaign Institute, Zhejiang University, Haining 314400, China (e-mail: haoyang@intl.zju.edu.cn)}
\thanks{W. Feng is with the Department of Electronic Engineering, State Key Laboratory of Space Network and Communications, Tsinghua University, Beijing 100084, China (e-mail: fengwei@tsinghua.edu.cn) \textit{(Corresponding author: Wei Feng.)} }
\thanks{T. Q. S. Quek is with the Information Systems Technology
and Design Pillar, Singapore University of Technology and Design, Singapore 487372 (e-mail: tonyquek@sutd.edu.sg).}
}

\maketitle

\begin{abstract}
Federated learning (FL) and split learning (SL) are two effective distributed learning paradigms in wireless networks, enabling collaborative model training across mobile devices without sharing raw data. While FL supports low-latency parallel training, it may converge to less accurate model. In contrast, SL achieves higher accuracy through sequential training but suffers from increased delay. To leverage the advantages of both, hybrid split and federated learning (HSFL) allows some devices to operate in FL mode and others in SL mode. This paper aims to accelerate HSFL by addressing three key questions: 1) How does learning mode selection affect overall learning performance? 2) How does it interact with batch size? 3) How can these hyperparameters be jointly optimized alongside communication and computational resources to reduce overall learning delay? We first analyze convergence, revealing the interplay between learning mode and batch size. Next, we formulate a delay minimization problem and propose a two-stage solution: a block coordinate descent method for a relaxed problem to obtain a locally optimal solution, followed by a rounding algorithm to recover integer batch sizes with near-optimal performance. Experimental results demonstrate that our approach significantly accelerates convergence to the target accuracy compared to existing methods.
\end{abstract}


\pagestyle{empty}  
\thispagestyle{empty} 

\section{Introduction}

\begin{figure}[t]
\centering
\subfloat{
		\includegraphics[width=8.8cm]{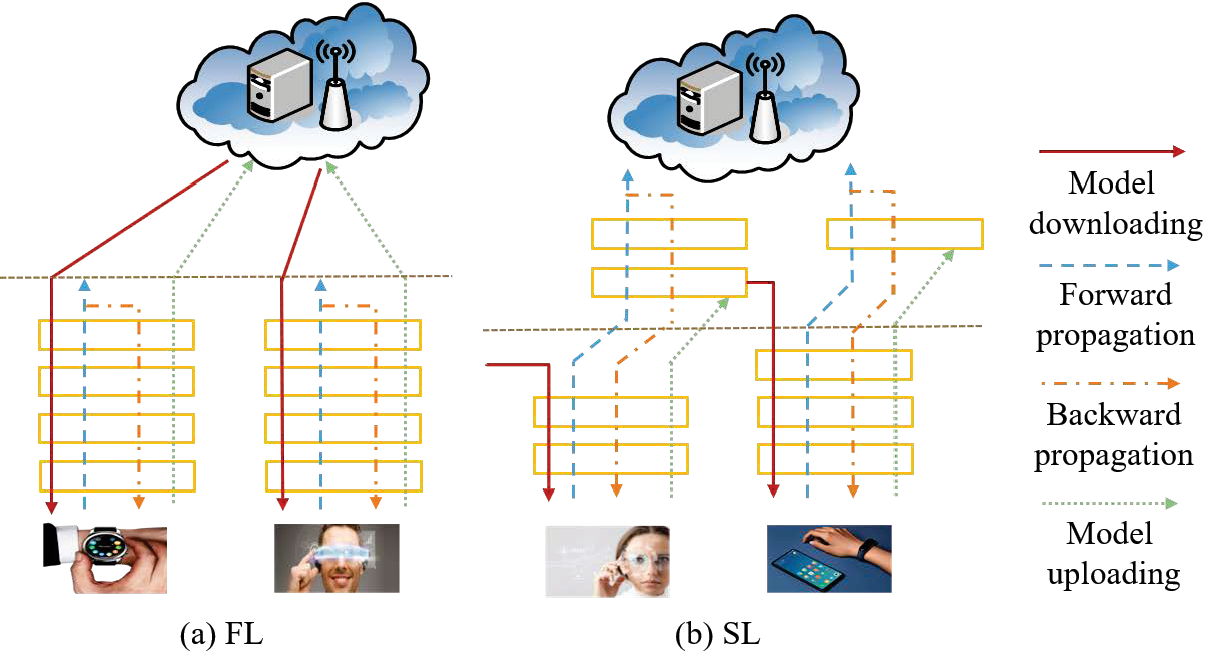}}
\caption{An illustration of distributed learning workflow, where the rectangle represents the model layer: (a) FL workflow; (b) SL workflow.}
\label{fig:Architecture}
\end{figure}

The rapid proliferation of mobile devices has led to an explosion in data generation, fueling the development of wireless distributed intelligence. In this context, federated learning (FL) and split learning (SL), two mainstream paradigms in distributed learning, have emerged as promising solutions in wireless networks \cite{IEEEComSoc2025AI,3gppTR22874}. As shown in Fig. \ref{fig:Architecture}, FL enables mobile devices to collaboratively train a global deep neural network (DNN) model without exposing their raw data \cite{Google_FL}. Each device downloads the global model from an edge server, performs local training using its private data, and uploads model updates to the server for global model refinement. 
This process is repeated over several communication rounds until convergence. Alternatively, SL partitions the DNN model into two segments: a local model and an edge model, which are updated on the device and edge server, respectively. Originally proposed in \cite{SL}, SL adopts a sequential training scheme where devices interact with the server one at a time, performing collaborative forward and backward propagation to update the respective model components.

While FL supports parallel training across mobile devices, it requires each device to locally train the entire model, which overwhelms those with limited computational capabilities. Furthermore, the presence of non-independent and identically distributed (non-IID) data across devices, also referred to as data heterogeneity, degrades convergence speed and model accuracy, often leading to a biased global model if not properly addressed \cite{Kun_JSTSP,ICML_Junyu}. In contrast, SL alleviates the computational burden on devices by offloading a portion of the model to the server and its sequential training process resembles centralized learning, thereby mitigating the adverse effects of data heterogeneity \cite{Centralized_EdgeLearning,li2023convergence}. However, this sequential nature introduces significant learning delay, particularly in scenarios involving a large number of devices. Therefore, 
FL enables low-latency training but may converge to less accurate model, while SL achieves higher accuracy at the cost of increased learning delay. These trade-offs motivate the exploration of hybrid approaches that leverage the strengths of both paradigms.

Naturally, hybrid split and federated learning (HSFL) has emerged \cite{liu2022wireless}, allowing mobile devices to engage in distributed learning using either FL or SL mode. In this framework, a subset of devices performs SL while the remaining devices execute FL, both operating in parallel. Compared to pure SL, reducing the number of SL participants helps lower the overall learning delay. At the same time, the inclusion of SL enhances learning performance over pure FL, particularly in heterogeneous data environments. However, HSFL introduces several key challenges: 
\begin{itemize}
    \item \textbf{Q1:} What effect does learning mode selection have on overall learning performance?
    \item \textbf{Q2:} How does the newly introduced hyperparameter, i.e., learning mode selection, interact with conventional ones, such as batch size?
    \item \textbf{Q3:} How can these hyperparameters be jointly optimized, along with communication and computational resource management, to accelerate HSFL in resource-constrained wireless networks?
\end{itemize}

To address these questions, we first conduct a convergence analysis of HSFL to reveal the impact of learning mode selection and batch size on the learning performance. Based on these theoretical insights, we formulate an optimization problem to minimize the overall learning delay, by jointly optimizing hyperparameter configurations (learning mode selection and batch size) and network resource management (model splitting and bandwidth allocation). Leveraging the problem structure, we adopt the block coordinate descent method to obtain a locally optimal solution under relaxed batch size constraint. Furthermore, we propose a batch size rounding algorithm to restore integer-valued batch sizes while preserving near-optimal performance. Experimental results validate the effectiveness of our proposed algorithm, showing that it quickly achieves the target accuracy with improved learning efficiency.

The main contributions of this work lie in addressing questions Q1-Q3, as summarized below:
\begin{itemize}
    \item \textbf{A1:} Through convergence analysis, we deduce that involving more devices in SL reduces the number of rounds required for convergence. However, due to the sequential nature of SL, this comes at the cost of increased learning delay in each round.
    \item \textbf{A2:} Increasing the batch size enables more data samples to be used in model training, which decreases the number of rounds to convergence but increases one-round delay. Given their consistent impact on learning performance, a tradeoff arises between the number of devices participating in SL and the batch size.
    \item \textbf{A3:} Owing to the trade-off between one-round delay and the number of rounds to convergence, there exists an optimal combination of learning mode selection and batch size. To minimize the overall learning delay, we further jointly optimize model splitting for SL and bandwidth allocation between SL and FL, which are two key factors influencing one-round delay under limited wireless network resources. 
\end{itemize}

The rest of this paper is organized as follows: We review related works in Section \ref{sec:related_works} and elaborate on the learning flow and delay model in Section \ref{sec:learning_flow}. In Section \ref{sec:problem_formulation}, we present the convergence analysis and problem formulation, followed by algorithm design in Section \ref{sec:algorithm_design}. Experimental results are given in Section \ref{sec:exp_results} and conclusions are drawn in Section \ref{sec:conclusions}.

\section{Related Works \label{sec:related_works}}
In this section, we review the related literature from two perspectives: wireless distributed learning frameworks that integrate SL and FL, and optimization methods designed to further enhance learning performance.

\subsection{Wireless Distributed Learning Frameworks}
To harness the advantages of both FL and SL, existing studies have explored their integration. SplitFed, also referred to as SFL in the literature, was first introduced in \cite{SplitFed} as an FL framework that incorporates model splitting, where a portion of the model is trained locally on devices and the remainder is processed on the server. To enhance SplitFed's performance, activation aggregation at the cut layer (approximating features derived from IID data) and at the last layer (reducing the dimensionality of activation gradients) were proposed in \cite{mergesfl} and \cite{Linzheng}, respectively, aiming to improve model accuracy and reduce learning delay. \cite{GAN_SFL} explored the use of  delayed gradients for multiple local updates within the SplitFed framework, enabling communication-computation parallelism to reduce both learning delay and energy consumption with minimal or no compromise in model accuracy. Additionally, \cite{li2024introducing} and \cite{local_loss} employed local model training to enhance personalization, while leveraging full model training to improve generalization. A common theme among these works is the use of model splitting, as in SL, to alleviate the computational burden on devices encountered in FL. However, they primarily focus on a single learning mode, in contrast to our work, which considers the coexistence of multiple learning modes.

Building on the sequential SL, \cite{wu2023split} proposed a cluster-based SL framework, where devices are grouped into clusters for parallel intra-cluster training and sequential inter-cluster training. \cite{liu2022wireless} introduced HSFL, in which a subset of devices participates in FL while others engage in SL. Both approaches exploit the sequential nature of SL to enhance model accuracy. Compared to \cite{wu2023split}, HSFL provides greater flexibility in learning mode selection, offering the potential for further performance gains. Although \cite{wu2023split} initially demonstrated the promise of HSFL, key questions Q1-Q3 remain unsolved.

\subsection{Learning Optimization Methods}
A common approach for enhancing learning performance in wireless networks is the joint optimization of learning hyperparameters and network resources. A fundamental challenge in this area is making the implicit impact of hyperparameters on learning performance explicit and analytically tractable, which often depends on the specific learning paradigm. In what follows, we review related works on FL, SL, and their integrated paradigms, respectively.

In SL, model splitting significantly affects learning delay while generally having no impact on model accuracy. Based on this insight, \cite{Yuzhu_ICCC} investigated dynamic wireless environments and employed deep reinforcement learning to adaptively optimize the model splitting for minimizing the overall learning cost. Extending this line of work, \cite{wang2021hivemind} considered a more complex SL setting with multiple possible cut layers and proposed a low-complexity algorithm for selecting the optimal cut-layer configuration.

In FL, key factors such as device scheduling \cite{Semi_chaoqun,mobility_neighbours}, aggregation frequency \cite{FL_shiqiang,dynamite}, and batch size \cite{totalkortowork,adaptivebatchsize} play critical roles in learning efficiency. Since our work focuses on batch size optimization, we highlight several representative studies in this area. \cite{dynamite} jointly optimized batch size and aggregation frequency to balance convergence rate, system cost, and completion time. \cite{totalkortowork} optimized batch sizes across rounds, with an increasing coefficient derived from convergence analysis. 
\cite{adaptivebatchsize} analyzed the coupling between batch size and learning rate, and adaptively adjusted batch sizes with scaled learning rate to synchronize the one-round delay across devices.  \cite{rethikingresource} proposed a two-phase training framework combining general pre-training with task-specific fine-tuning, and conducted convergence analysis to guide the joint optimization of batch sizes and network resources.

In integrated paradigms such as SplitFed, batch size optimization has drawn increasing attention due to its impact on both computation and communication overhead. For example, \cite{mergesfl} optimized batch sizes to reduce device waiting time in each round, but did not consider their impact on the number of rounds to convergence. In the context of HSFL, a newly introduced hyperparameter, i.e., learning mode selection, leads to distinct convergence behavior. Therefore, the batch size optimization methods proposed in prior works such as \cite{dynamite,adaptivebatchsize,totalkortowork,mergesfl,rethikingresource} are not directly applicable to HSFL. Although the conference version of this work \cite{GC_Xuefei} preliminarily explored the learning mode selection problem in HSFL, the interaction between learning mode selection and batch size optimization remains unaddressed. This open issue motivates the further investigation presented in this paper.

\section{Learning Workflow and Delay Model\label{sec:learning_flow}}

In this section, we first introduce the learning workflow to illustrate the HSFL process, followed by a detailed explanation of the delay model.

\subsection{Learning Workflow\label{sub:learning_flow}}
We consider a server and $K$ mobile devices for wireless distributed learning. The dataset of device $k\in\mathcal{K} \triangleq \left\{1,...,K\right\}$ is denoted by $\mathcal{D}_k$, including $D_k$ data samples. In detail, data sample $i$ is expressed as $\{\mathbf{x}_k^i,y_k^i\}$, where ${\mathbf{x}}_k^i$ and $y_k^i$ are the input data point and the label for DNN model training, respectively. The learning goal is to find the optimal model $\boldsymbol{\omega}^{*}$ with minimum global loss, as follows:
\begin{equation}
   \boldsymbol{\omega}^* = \arg \min_{\boldsymbol{\omega}} ~L(\boldsymbol{\omega}).    
\end{equation}
More specific, the global loss function is given by
\begin{equation}
\begin{aligned}
L(\boldsymbol{\omega})  = \frac{1}
{K}\sum_{k=1}^K  L_k(\boldsymbol{\omega},\mathcal{D}_k),
\end{aligned}
\end{equation}
with local loss function defined as
\begin{equation}
\begin{aligned}
L_k(\boldsymbol{\omega},\mathcal{D}_k) = \frac{1}{
D_{k}} \overset{ D_k }{\underset{i=1}{\sum}} l\left(\boldsymbol{\omega},\mathbf{x}_k^i,y_k^i\right),
\end{aligned}
\end{equation}
where $l(\boldsymbol{\omega},\mathbf{x}_k^i,y_k^i)$ is the loss function computed with data sample $i$ of device $k$.

In a distributed manner, HSFL approaches $\boldsymbol{\omega}^*$ with multiple communication rounds. In round $t$, each device chooses to participate in either FL or SL, referred to as an FL device or an SL device, respectively. Let $\mathcal{K}_t^{\rm{F}}$ denote the set of $K_t^{\rm{F}}$ FL devices and $\mathcal{K}_t^{\rm{S}}$ denote the set of $K_t^{\rm{S}}$ SL devices. Below, we elaborate on the three key procedures in HSFL.

\subsubsection{Model Training of FL Devices} At the beginning of round $t$, the server broadcasts global model $\boldsymbol{\omega}_t$ to all FL devices. Then, FL devices perform local training in parallel and, upon completion, upload model update to the server via frequency division multiple access (FDMA). The local model of FL device $k$ is updated as follows: 
\begin{equation}
    \boldsymbol{\omega}_{k,t+1} = \boldsymbol{\omega}_t - \eta_t \boldsymbol{g}_{k,t}(\boldsymbol{\omega}_t), \forall k\in\mathcal{K}_t^{\rm{F}},
    \label{eq:FL_GD}
\end{equation}
where $\eta_t$ is the learning rate and local gradient is given by
\begin{equation}
    \boldsymbol{g}_{k,t} (\boldsymbol{\omega}_t) = \nabla L_k(\boldsymbol{\omega}_t,\Xi_{k,t}).
    \label{eq:gradient}
\end{equation}
In \eqref{eq:gradient}, $\Xi_{k,t}$ denotes a mini-batch of size $\xi_{k,t}$, randomly sampled from the local dataset $\mathcal{D}_k$. The mini-batch stochastic gradient $\nabla L_k(\boldsymbol{\omega}_t,\Xi_{k,t})$ provides an unbiased estimate of the full gradient $\nabla L_k(\boldsymbol{\omega}_t,\mathcal{D}_{k})$, while reducing local training cost.

\subsubsection{Model Training of SL Devices} For SL devices, model training is performed in a sequential manner. At the beginning of round $t$, the server randomizes the execution order of SL devices and determines a model cut-layer for each. The first SL device uses the global model $\boldsymbol{\omega}_t$ as its initial model. Each subsequent SL device, in turn, adopts the model update from its predecessor, i.e., the $k+1$-th SL device uses the model update from the $k$-th device as its initial model. As illustrated in Fig. \ref{fig:Architecture}, the initial model is partitioned into a local model and an edge model based on the selected cut-layer. The training process for the $k$-th SL device consists of the following steps: 
\begin{itemize}
    \item Downloads the local model from the server;
    \item Performs the forward and backward propagation successively, during which the activations and gradients \textit{w.r.t.} the model cut-layer, as well as the sample labels in the batch are exchanged between the $k$-th SL device and the server over a dedicated channel (orthogonal to those used by FL devices) \cite{OFDM_ChaoXu}; 
    \item Sends the local model update to the server, which then combines it with the corresponding edge model update to form a complete initial model for $k+1$-th SL device.
\end{itemize}
Hence, the model update rule for SL devices is given by 
\begin{equation}
\label{eq: SL_GD}
\begin{aligned}
    &\boldsymbol{\omega}_{1,t+1} = \boldsymbol{\omega}_t  -\eta_t \boldsymbol{g}_{1,t}\left(\boldsymbol{\omega}_t\right),\\
	&\boldsymbol{\omega}_{2,t+1} = \boldsymbol{\omega}_{1,t+1}  -\eta_t \boldsymbol{g}_{2,t}\left(\boldsymbol{\omega}_{1,t+1}\right),...,\\
    &\boldsymbol{\omega}_{K^{\rm{S}}_t,t+1}=\boldsymbol{\omega}_{K_t^{\rm{S}}-1,t+1} -\eta_t \boldsymbol{g}_{K^{\rm{S}}_t,t}\left(\boldsymbol{\omega}_{K_t^{\rm{S}}-1,t+1}\right),
\end{aligned}
\end{equation}
where $\boldsymbol{g}_{k,t}(\cdot)$ is calculated using a random batch $\Xi_{k,t}$ sampled from the local dataset $\mathcal{D}_k$, consistent with \eqref{eq:gradient}.

\subsubsection{Model Aggregation of FL and SL Devices} Upon receiving the model updates of all FL and SL devices, the server aggregates them to update the global model as follows: 
\begin{equation}
    \label{e2}
	\boldsymbol{\omega}_{t+1} = \frac{1}{K} \sum_{k=1}^{K} \boldsymbol{\omega}_{k,t+1}.
\end{equation}

\subsection{Learning Delay}
In each communication round, the learning delay is determined by the slower of the FL or SL training processes, due to synchronous model aggregation at the server. Excluding the negligible model aggregation delay, the learning delay in round $t$ is calculated as
\begin{equation}
\begin{aligned}
      T_t = 
    \max\{T_t^{\mathrm{F}},T_t^{\mathrm{S}}\},
\end{aligned}
\label{eq:round_delay}
\end{equation}
where $T_t^{\mathrm{F}}$ and $T_t^{\mathrm{S}}$ are the FL delay and SL delay respectively, indicating the time it takes from when the server sends the global model to the FL and SL devices until it receives their model updates. The detailed calculations of these delays are provided below, with all delay components considered within round $t$, unless otherwise specified. 

\subsubsection{FL Delay} 
Since all FL devices train and upload their local model in parallel, the FL delay depends on the lowest FL device, as follows:
\begin{equation}
    T_t^{\mathrm{F}} =  \underset{{k \in \mathcal K^{\mathrm F}_t}}{\mathrm{max}} \{\underbrace{ d_{k,t}^{\mathrm{FMD}} +d_{k,t}^{\mathrm{FMT}}+d_{k,t}^{\mathrm{FMU}}}_{T_{k,t}^{\rm{F}}}\}.
    \label{eq:FL_delay}
\end{equation}
For FL device $k$, its learning delay $T_{k,t}^{\rm{F}}$ is the sum of three key components, defined as
\begin{itemize}
    \item Model download delay in FL: The server broadcasts the global model to all FL devices at the following transmission rate: 
    \begin{equation}
    \label{eq1}
    R_{0,t}=\underset{k\in \mathcal{K}_t^{\text{F}}}{\text{min}}\left\{B_0\mathrm{log}_2\left(1+\frac{p_{0} h^{\rm{B}}_{k,t}}{\sigma B_0}\right)\right\},
    \end{equation}
    where $B_0$, $p_{0}$, $\sigma$, and $h_{k,t}^{\rm{B}}$ represent the broadcast channel bandwidth, the server transmit power, the noise power spectral density, and the broadcast channel gain between the server and FL device $k$. Consequently, the model download delay for FL device $k$ is given by 
    \begin{equation}
    \label{eq1}
    d_{k,t}^{\mathrm{FMD}}=\frac{S}{R_{0,t}},\forall k \in \mathcal K^{\mathrm F}_t,
    \end{equation}
    with $S$ representing the number of bits required to transmit the global model.
    \item Model training delay in FL: Upon receiving the global model, FL device $k$ begins training its local model using a randomly selected data batch of size $\xi_{k,t}$, resulting in the following delay:  
    \begin{equation}
    \label{eq1}
    	d_{k,t}^{\mathrm{FMT}}=\frac{\xi_{k,t} C}{f_{k,t}},\forall k\in \mathcal K^{\mathrm F}_t,
    \end{equation}
    where $f_{k,t}$ in floating point operations per second (FLOPs/s) denotes the computing capability of device $k$ and $C$ represents the number of FLOPs required to process a single data sample during both forward and backward propagation.
    \item Model upload delay in FL: After completing local model training, FL devices use FDMA to upload their model updates to the server for aggregation. Specifically, the model upload delay for FL device $k$ is given by 
    \begin{equation}
    \label{eq1}
    	d_{k,t}^{ {\mathrm{FMU}}}=\frac{S}{R_{k,t}^\mathrm{FU}},\forall k\in \mathcal K^{\mathrm {F}}_t,
    \end{equation}  
    where $R_{k,t}^\mathrm{FU}$ denotes the uplink transmission rate, calculated as 
    \begin{equation}
        \label{eq1}
        R_{k,t}^\mathrm{FU} = b_{k,t} B\mathrm{log}_2\left(1+\frac{p_{k} h^{\text{U}}_{k,t}}{\sigma b_{k,t} B}\right),\forall k \in \mathcal K^{\mathrm F}_t,
    \end{equation}
    with $p_{k}$ representing the transmit power of device $k$, $h_{k,t}^{\text{U}}$ the uplink channel gain between FL device $k$ and the server, $B$ the total system bandwidth,  and $b_{k,t}$ the bandwidth allocation ratio for FL device $k$. 
    \end{itemize}

\subsubsection{SL Delay}
Since SL devices train their models sequentially, the SL delay is computed as the total learning delay across all SL devices:
\begin{equation}
\begin{aligned}
    T_t^{\mathrm{S}}= {\underset{{k\in\mathcal{K}_t^{\rm{S}}}}{\sum}}( \underbrace{d_{k,t}^{\mathrm {SMD}}+d_{k,t}^{\mathrm{SMT}}+d_{k,t}^{\mathrm{SMU}}}_{T_{k,t}^{\rm{S}}}).
\end{aligned}
\label{eq:SL_delay}
\end{equation}
For SL device $k$, its learning delay $T_{k,t}^{\rm{S}}$ also consists of three key components. Before detailing these components, we first introduce the concept of model splitting. Assume that the trained DNN model comprises $L$ logical layers, each containing one or more physical layers. A partition at layer $l_{k,t}\in\mathcal{L}\triangleq \{1,...,L\}$ indicates that local model, consisting of layers 1 to $l_{k,t}$, is trained on device $k$, while the edge model, consisting of layers $l_{k,t}+1$ to $L$, is trained on the server. 
Based on this cut-layer $l_{k,t}$, the three key delay components are outlined below: 

\begin{itemize}
    \item Model download delay in SL: SL device $k$ downloads the latest local model from the server with the downlink transmission rate as follows:
    \begin{equation}
    \label{eq1}
    R_{k,t}^\mathrm{SD} = b_{0,t} B\mathrm{log}_2\left(1+\frac{p_{0} h^{\rm{D}}_{k,t}}{\sigma b_{0,t} B}\right),\forall k \in \mathcal K^{\mathrm S}_t,
    \end{equation}
    where $h_{k,t}^{\rm{D}}$ is the downlink channel gain between the server and device $k$. In addition, $b_{0,t}$ is the bandwidth allocation ratio dedicated to the SL training process, which is sequentially shared among all SL devices. The model download delay for SL device $k$ is given by
    \begin{equation}
    \label{eq:download_delay_SL}
    	d_{k,t}^{ {\mathrm{SMD}}}=\overset{l_{k,t}}{\underset{l=1}{\sum}}\frac{s_l}{R_{k,t}^\mathrm{SD}},\forall k\in \mathcal K^{\mathrm S}_t,
    \end{equation}
    where $s_l$ is the number of bits required to transmit model parameters at layer $l$.
    \item Model training delay in SL: For SL device $k$, its model training delay is given by 
    \begin{equation}
        d_{k,t}^{\rm{SMT}} = d_{k,t}^{\rm{CMP}}+d_{k,t}^{\rm{COM}}, \forall k\in \mathcal K^{\mathrm S}_t,
        \label{eq:training_delay_SL}
    \end{equation}
    where $d_{k,t}^{\rm{CMP}}$ denotes the computation delay for both local and edge model training, and is calculated as 
    \begin{equation}
        \label{eq1}   d_{k,t}^{\mathrm{CMP}}=\xi_{k,t}\left(\overset{l_{k,t}}{\underset{l=1}{\sum}}\frac{ c_l}{f_{k,t}} + \underset{l=l_{k,t}+1}{\overset{L}{\sum}}\frac{ c_l}{f_{0,t}}\right), \forall k\in \mathcal K^{\mathrm S}_t,
    \end{equation}
    where $c_l$ denotes the number of FLOPs required to train model parameters at layer $l$ using a single data sample, and $f_{0,t}$ in FLOPs/s represents the computing capability of the server. Additionally, $d_{k,t}^{\rm{COM}}$ is the communication delay incurred during the transmission of activations, labels, and gradients between SL device $k$ and the server, given by 
    \begin{equation}
    \label{eq1}
    	d_{k,t}^{ {\mathrm{COM}}}=\xi_{k,t}\left(\frac{o_{l_{k,t}}^{\rm{F}}}{R_{k,t}^\mathrm{SU}}+\frac{o_{l_{k,t}}^{\rm{B}}}{R_{k,t}^\mathrm{SD}}\right),\forall k\in \mathcal K^{\mathrm S}_t,
    \end{equation}
    where $o_{l_{k,t}}^{\rm{F}}$ is the number of bits required to transmit the activations at cut-layer $l_{k,t}$ along with the label, during the forward propagation of a single data sample. Besides, $o_{l_{k,t}}^{\rm{B}}$ is the number of bits needed to transmit the gradients \textit{w.r.t.} the activations at cut-layer $l_{k,t}$, during the backward propagation of one sample. Considering time-division duplex for uplink and downlink transmission between SL device $k$ and the server, we express the uplink transmission rate as 
    \begin{equation}
    \label{eq1}
    	R_{k,t}^\mathrm{SU} = b_{0,t} B\mathrm{log}_2\left(1+\frac{p_{k} h^{\text{U}}_{k,t}}{\sigma b_{0,t} B}\right),\forall k \in \mathcal K^{\mathrm S}_t,
    \end{equation}
    where $h_{k,t}^{\rm{U}}$ is the uplink channel gain between SL device $k$ and the server.
    \item Model upload delay in SL: After completing model training, SL device $k$ uploads its local mode update to the server, resulting in the following delay: 
    \begin{equation}
    \label{eq1}
    	d_{k,t}^{ {\mathrm{FMU}}}=\overset{l_{k,t}}{\underset{l=1}{\sum}}\frac{s_l}{R_{k,t}^\mathrm{SU}},\forall k\in \mathcal K^{\mathrm S}_t.
    \end{equation}
 \end{itemize}

From the above calculations, we find that two key hyperparameters, i.e., learning mode selection and batch size, have a significant impact on the learning delay. In addition, bandwidth allocation and model splitting also influence the delay performance. Therefore, we proceed to optimize these factors to accelerate the HSFL process. 
 
\section{Convergence Analysis and Problem Formulation\label{sec:problem_formulation}}

In this section, we first conduct the convergence analysis of HSFL to reveal the impact of learning mode selection and batch size on the learning performance, followed by a problem formulation to optimize HSFL for low-latency and high-accuracy wireless distributed intelligence.

\subsection{Convergence Analysis}
We make the following assumptions on the loss function and the stochastic gradients \cite{bottou2018optimization}:

\newtheorem{assumption}{Assumption}
\begin{assumption}
\label{ass:lipschitz}
$\kappa$-Lipschitz continuity: The global loss function satisfies: $\| L(\boldsymbol{\omega})-L(\boldsymbol{v}) \|\leq \kappa \|\boldsymbol{\omega}-\boldsymbol{v} \|$ for all $\boldsymbol{v} \text{ and }  \boldsymbol{\omega}$, where $\kappa>0$ is a constant. 
\end{assumption}

\begin{assumption}
\label{ass:lsmooth}
$l$-smoothness: For each device $k$, its local loss function satisfies: $L_k(\boldsymbol{\upsilon},\mathcal{D}_k)\leq L_k(\boldsymbol{\omega},\mathcal{D}_k)+\langle \boldsymbol{\upsilon}-\boldsymbol{\omega},\nabla L_k(\boldsymbol{\omega},\mathcal{D}_k)\rangle+\frac{\ell}{2}\left\|\boldsymbol{\upsilon}-\boldsymbol{\omega}\right\|_2^2$ for all $\boldsymbol{v} \text{ and } \boldsymbol{\omega}$, where $\ell>0$ is a constant. 
\end{assumption}

\begin{assumption}
\label{ass:secondMoment}
Second moment bounds: For any given model $\boldsymbol{w}$ and mini-batch $\Xi$, the following conditions hold for each device $k$: 
\begin{equation}
    \begin{aligned}
        \mathbb{E} \left[||\nabla L_{k}(\boldsymbol{\omega},\Xi)||^2\right] & \le G^2\\
        \mathbb{V}[\nabla l(\boldsymbol{\omega},\mathbf{x}_k^i,y_k^i)] & \leq M^2,
    \end{aligned}
\end{equation}
indicating that both the expected squared norm of the mini-batch stochastic gradient and the variance of the gradient computed on any single data sample $(\mathbf{x}_k^i,y_k^i)\in\mathcal{D}_k$ are bounded, where $G^2$ and $M^2$ as positive constants.
\end{assumption}


By conducting a convergence analysis, we derive Theorem 1 to reveal the impact of both learning mode selection and batch size on learning performance.
\begin{thm}
   Based on Assumptions \ref{ass:lipschitz},\ref{ass:lsmooth}, and \ref{ass:secondMoment}, we have
    \begin{equation} \mathbb{E}\left[\left\|L(\boldsymbol{\omega}_{t+1})-L(\boldsymbol{\omega}^*)\right\|^2\right] \leq  \gamma_1\mathbb{E}\left[\left\|\boldsymbol{\omega}_t-\boldsymbol{\omega}^*\right\|^2\right]+  W_t,
    \end{equation}
    where $W_t$ is expanded as
    \begin{equation}
        W_t=\frac{\gamma_2}{K}\sum_{k=1}^K \frac{1}{\xi_{k,t}} +\gamma_3\left(K-\frac{K_t^{\rm{S}}(K_t^{\rm{S}}-1)}{2K}\right) +\gamma_4\Phi.
    \end{equation}
    Note that, $\gamma_1$, $\gamma_2$, $\gamma_3$, and $\gamma_4$ are positive constants that depend on $\kappa$, $\ell$, $G^2$, $M^2$, and learning rate $\eta_t$. Moreover, $\Phi$ is a constant that captures the heterogeneity of the data distribution among devices, with smaller values indicating lower heterogeneity levels.
\end{thm}
\proof The proof is given in the appendices, which are provided as supplementary material due to space limitations. \qed

\begin{rem}
    From Theorem 1, we derive the following insights to guide the key hyperparameter settings in each round $t$:
\begin{itemize}
    \item Increasing batch size $\xi_{k,t}$ for model update of any device $k$ is beneficial to reduce the first term in $W_t$, therefore moving $\boldsymbol{\omega}_{t+1}$ closer to the optimal model $\boldsymbol{\omega}^*$.
    \item Increasing $K_t^{\rm{S}}(K_t^{\rm{S}}-1)$ reduces the second term in $W_t$, facilitating faster convergence toward $\boldsymbol{\omega}^*$. Notably, when $K_t^{\rm{S}}=0$ holds, the HSFL reduces to standard FL; and when $K_t^{\rm{S}}=1$ holds, only a single SL device participates in the HSFL. As explained in the learning workflow in Section \ref{sub:learning_flow}, the global model updates in these two special cases are identical, resulting in equivalent convergence behavior. Therefore, increasing  $K_t^{\rm{S}}(K_t^{\rm{S}}-1)$ is effectively equivalent to increasing $K_t^{\rm{S}}$, making it a critical factor for HSFL acceleration.
    \item There exists a tradeoff between increasing $\xi_{k,t}$ and increasing $K_t^{\rm{S}}$. Specifically, to achieve the same loss function value, one may either use a larger $\xi_{k,t}$ with a smaller $K_t^{\rm{S}}$, or a smaller $\xi_{k,t}$ with a larger $K_t^{\rm{S}}$.
    \item An increase in $\xi_{k,t}$ or $K_t^{\rm{S}}$ can effectively reduce the number of rounds needed to converge toward $\boldsymbol{\omega}^*$.
\end{itemize}
\end{rem}

Inspired by these insights, we proceed to optimize the learning mode selection and batch size, under the constraints of limited communication and computational resources in wireless environments. 

\subsection{Problem Formulation}
Although increasing $K_t^{\rm{S}}$ and $\xi_{k,t}$ can reduce the number of rounds for learning convergence, they also increase the per-round delay in resource-restricted wireless environments. Since both the number of rounds and the per-round delay critically impact the overall learning delay, it is essential to strike a balance between them to accelerate the learning. Therefore, in round $t$, we aim to minimize the following objective function:
\begin{equation}
u_t(\bm{x}_{t},\bm{l}_{t},\bm{b}_{t}, \bm{\xi}_{t}) = T_t
 - \rho_1 K_t^{\rm{S}}(K_t^{\rm{S}}-1) +\sum_{k=1}^K \frac{\rho_2}{\xi_{k,t}},\\
\label{eq:obj}
\end{equation}
where $T_t$ is the one-round delay, as defined in \eqref{eq:round_delay}, while the subsequent two terms are derived from Theorem 1 to capture the impact of $K_t^{\rm{S}}$ and $\xi_{k,t}$ on the total number of rounds. The coefficients $\rho_1>0$ and $\rho_2>0$ serve as weighting factors. We define the learning mode selection and batch size variables as $\bm{x}_t=\{x_{k,t}|k\in\mathcal{K}\}$ and $\bm{\xi}_t=\{\xi_{k,t}|k\in\mathcal{K}\}$. Specifically, $x_{k,t}\in\{0,1\}$ indicates the learning mode of device $k$, where $x_{k,t} = 0$ corresponds to FL mode and $x_{k,t} =1$ to SL mode. Accordingly, the sets of FL and SL devices are defined as $\mathcal{K}_t^{\rm{F}}=\{k|x_{k,t}= 0, k\in\mathcal{K}\}$ and $\mathcal{K}_t^{\rm{S}}=\{k|x_{k,t}= 1, k\in\mathcal{K}\}$, respectively. To efficiently utilize the limited communication and communicational resources, we also optimize the model splitting and bandwidth allocation, denoted by $\bm{l}_t=\{l_{k,t}|k\in\mathcal{K}_t^{\rm{S}}\}$ and $\bm{b}_t = \{b_{k,t},b_{0,t}|k\in\mathcal{K}_t^{\rm{F}}\}$. 

Furthermore, the optimal problem is formulated as
\begin{equation}
\begin{aligned} 
    {\text{(P0) }} \min_{\bm{x}_{t},\bm{l}_{t},\bm{b}_{t},\bm{\xi}_{t}} & \; u_t(\bm{x}_{t},\bm{l}_{t},\bm{b}_{t},\bm{\xi}_{t})\nonumber \\
    \text{s.t.} \quad\,
    &  ~\text{C1:} \; x_{k,t} \in \{0,1\},\forall k \in \mathcal K \nonumber\\  
    &   ~\text{C2:} \; l_{k,t} \in \mathcal L,\forall k \in \mathcal K_t^{\mathrm{S}} \nonumber \\
    & ~\text{C3:} \; \sum_{k\in \mathcal K_t^{\text{F}} } b_{k,t} + b_{0,t} \le 1 \nonumber\\
    & ~\text{C4:} \; b_{0,t}\in [0,1]\\
    & ~\text{C5:} \; b_{k,t} \in [0,1]  ,\forall k \in \mathcal K^{\mathrm F}_t\\
    & ~\text{C6:} \; 1\le \xi_{k,t}\le D_{k},\forall k \in \mathcal K\\
    &~\text{C7:} \; \xi_{k,t}\in {\mathbb{Z}^+},\forall k \in \mathcal K,
\end{aligned}
\end{equation}
where C1 and C2 put restrictions on the mode selection of all devices and the model splitting for SL devices, respectively. C3-C5 regulate the bandwidth allocation for each device, while C6 and C7 define the feasible region for the batch size. In particular, $\mathbb{Z}^+$ in C7 denotes the set of positive integers. (P0) is a mixed integer programming problem. To improve its tractability, we next explore its structure properties and reformulate it to facilitate algorithm design.

\section{Algorithm Design\label{sec:algorithm_design}}
With the understanding that the block coordinate descent method can yield a locally optimal solution to the original problem when each subproblem over a block of variables is solved optimally \cite{BCD}, we tackle (P0) in two steps: 1) Treating $\{\bm{x}_t,\bm{l}_t,\bm{b}_t\}$ and $\bm{\xi}_t$ as two variable blocks, achieve a locally optimal solution to (P0) regardless of C7; 2) Round the resulting $\bm{\xi}_t$ to obtain a near-optimal integer solution that satisfies C7. Accordingly, we present the algorithm sketch, followed by block-wise optimization, and finally, the design of the batch size rounding algorithm in this section. 


\subsection{Algorithm Sketch}

Omitting C7, we decompose (P0) into the following two subproblems. One concerns the variable block $\{\bm{x}_{t},\bm{l}_{t},\bm{b}_{t}\}$:
\begin{equation}
\begin{aligned} 
    {\text{(P1) }}  \min_{\bm{x}_{t},\bm{l}_{t},\bm{b}_{t}} &~ u_{1t}(\bm{x}_{t},\bm{l}_{t},\bm{b}_{t})\nonumber \\
    \text{s.t.} \;\; &~\text{C1-C5},
\end{aligned}
\end{equation}
and the other one pertains to the variable block $\bm{\xi}_{t}$:
\begin{equation}
\begin{aligned} 
    {\text{(P2) }}  \min_{\bm{\xi}_{t}} &~ u_{2t}(\bm{\xi}_{t})\nonumber \\
    \text{s.t.} \,&~\text{C6},
\end{aligned}
\end{equation}
where the objective functions are given by 
\begin{equation}
\begin{aligned}
& u_{1t}(\bm{x}_{t},\bm{l}_{t},\bm{b}_{t}) =T_t
 - \rho_1 K_t^{\rm{S}}(K_t^{\rm{S}}-1) +\Gamma_1,\\
& u_{2t}(\bm{\xi}_{t}) = T_t
 +\sum_{k=1}^K \frac{\rho_2}{\xi_{k,t}} + \Gamma_2.\\
\end{aligned}
\label{eq: Obj_P1&P2}
\end{equation}
Note that, with fixed $\bm{\xi}_{t}$, the term $\Gamma_1= \sum_{k=1}^K \rho_2/\xi_{k,t}$ becomes a constant in (P1), while with fixed $\bm{x}_{t}$, the term $\Gamma_2=- \rho_1 K_t^{\rm{S}}(K_t^{\rm{S}}-1)$ is constant in (P2).

\begin{algorithm}[t] 
	\caption{Comprehensive Algorithm Sketch}
	\label{alg:alternating} 
	\begin{algorithmic}[1]
        \STATE Initialize $\bm{x}_{t}, \bm{l}_{t}, \bm{b}_{t}$, and $\bm{\xi}_t$, and calculate the initial objective function value of (P0) as $\hat{u}_t$; 
        \REPEAT
            \STATE Update $\mu_t \gets \hat{\mu}_t$;
            \STATE With fixed $\bm{\xi}_t$, solve problem (P1) using Algorithm~\ref{alg:Gibbs} to update $\{\bm{x}_{t}, \bm{l}_{t}, \bm{b}_{t}\}$;
            \STATE With fixed $\{\bm{x}_{t}, \bm{l}_{t}, \bm{b}_{t}\}$, solve problem (P2) using Algorithm~\ref{alg:BatchDecision} to update $\bm{\xi}_t$;
            \STATE Recalculate the objective function value as $\hat{u}_t$ based on the updated $\bm{x}_{t}, \bm{l}_{t}, \bm{b}_{t}$, and $\bm{\xi}_t$;
        \UNTIL{$u_t - \hat{u}_t \leq  \epsilon_1$}
        \STATE Round the resulting $\bm{\xi}_t$ using Algorithm~\ref{alg:BatchRound} to obtain integer batch sizes that satisfy C7;
        \STATE Based on the rounded $\bm{\xi}_t$, solve (P1) again using Algorithm~\ref{alg:Gibbs} to output a near-optimal solution to (P0).
	\end{algorithmic} 
\end{algorithm}
Next, we solve (P1) and (P2) alternately using the block coordinate descent method, as outlined in Steps 2-7 of Algorithm \ref{alg:alternating}. In each iteration, we first use Algorithm \ref{alg:Gibbs} to solve (P1), determining the optimal learning mode, model splitting, and bandwidth allocation decisions. We then design Algorithm \ref{alg:BatchDecision} to optimize (P2) and obtain the batch size. This alternating process continues until the change in objective value between successive iterations falls below a small positive threshold $\epsilon_1$, as shown in Step 7. Since C7 is not considered during the iterations, the resulting batch sizes may be continuous. Hence, we introduce a rounding procedure (Algorithm 6) to convert batch sizes to integers, yielding a near-optimal solution to (P0) as described in Steps 8 and 9. With this algorithm sketch in mind, we proceed to elaborate on Algorithms 4, 5, and 6 in sequence. 

\subsection{Learning Mode Selection, Model Splitting, and Bandwidth Allocation Optimization}

To effectively solve (P1), we reformulate it as
\begin{equation}
\begin{aligned} 
    {\text{(P3) }} \min_{\bm{x}_{t}} &~ u_{1t}^{*}(\bm{l}_{t},\bm{b}_{t}|\bm{x}_{t})\nonumber \\
    \text{s.t. } &~\text{C1},
\end{aligned}
\end{equation}
where $u_{1t}^*(\bm{l}_{t},\bm{b}_{t}|\bm{x}_{t})$ is the optimal value of the following problem, with $\bm{x}_{t}$ held fixed:
\begin{equation}
\begin{aligned} 
    {\text{(P4) }} \min_{\bm{l}_{t},\bm{b}_{t}}   ~&  u_{1t}(\bm{l}_{t},\bm{b}_{t}|\bm{x}_{t})\nonumber \\
    \text{s.t. } &~\text{C2-C5}.
\end{aligned}
\end{equation}
With this reformulation, we solve (P3) instead of directly tackling (P1) \cite{C-RAN-JSAC}. Specifically, we first solve (P4) to determine the model splitting and bandwidth allocation, and then substitute its optimal value into (P3) to select the learning mode.

\begin{algorithm}[t] 
	\caption{FL-device Bandwidth Allocation}
	\label{alg:FLBand} 
	\begin{algorithmic}[1]
	   \STATE Initialize $d_{\rm{L}} =        \max_{k\in\mathcal{K}_t^{\rm{F}}}\{ d_{k,t}^{\rm{FMD}}+d_{k,t}^{\rm{FMT}}\}$, and set \( d_{\rm{U}} \) as the objective value of (P7) under equal allocation of \( (1 - b_{0,t}) \) across all FL devices;
            \REPEAT
                \STATE Set $d^* = (d_{\rm{L}}+d_{\rm{U}})/2$;
                \STATE Compute $b_{k,t}^*$ by solving \eqref{eq:bandwidth_FL}; 
                \IF{$\sum_{k\in\mathcal{K}_t^{\rm{F}}}b_{k,t}^*<1-b_{0,t}-\epsilon_2$}
                    \STATE Update $d_{\rm{U}} = d^*$;
                \ELSIF{$\sum_{k\in\mathcal{K}_t^{\rm{F}}}b_{k,t}^*>1-b_{0,t}$}
                    \STATE Update $d_{\rm{L}} = d^*$;
                \ENDIF 
            \UNTIL{$1-b_{0,t}-\epsilon_2 \leq \sum_{k\in\mathcal{K}_t^{\rm{F}}} b_{k,t}^* \leq 1-b_{0,t}$} 
            \STATE Output $b_{k,t}^*$ as the optimal solution of (P7).
	\end{algorithmic} 
\end{algorithm}

\subsubsection{Model Splitting and Bandwidth Allocation Design}

We observe from (P4) that once $b_{0,t}$ is specified, $\bm{l}_{t}$ and $\hat{\bm{b}}_{t} = \{b_{k,t}|k\in\mathcal{K}_t^{\rm{F}}\}$ become decoupled. Therefore, (P4) can be reformulated as 
\begin{equation}
\begin{aligned} 
    {\text{(P5) }} \min_{b_{0,t}}   ~&  u_{1t}^{*}(\bm{l}_{t},\hat{\bm{b}}_{t}|\bm{x}_{t},b_{0,t})\nonumber \\
    \text{s.t. } &~\text{C4},
\end{aligned}
\end{equation}
where the objective function is expanded as 
\begin{equation}
\begin{aligned}
    & u_{1t}^*(\bm{l}_{t},\hat{\bm{b}}_{t}|\bm{x}_{t},b_{0,t}) \\
    & = \max\left\{T_t^{\rm{S}}(\bm{l}_{t}^*|\bm{x}_{t},b_{0,t}),T_t^{\rm{F}}(\hat{\bm{b}}_{t}^*|\bm{x}_{t},b_{0,t})\right\}+\Lambda + \Gamma_1.
\end{aligned} 
\label{eq: delay_P3}
\end{equation}
In addition to $\Gamma_1$, the term $\Lambda=-\rho_1 K_t^{\rm{S}}(K_t^{\rm{S}}-1)$ is a constant with $\bm{x}_{t}$ held fixed. Moreover, $\bm{l}_{t}^*=\{l_{k,t}^*, k\in\mathcal{K}_t^{\rm{S}}\}$ and $\hat{\bm{b}}_{t}^*=\{b_{k,t}^*, k\in\mathcal{K}_t^{\rm{F}}\}$ denote the optimal solutions of the following two independent problems:
\begin{equation}
\begin{aligned} 
    {\text{(P6) }} \min_{\bm{l}_{t}}   ~&  T_t^{\rm{S}}(\bm{l}_{t}|\bm{x}_{t},b_{0,t})\nonumber \\
    \text{s.t. } &~\text{C2},\\
    {\text{(P7) }} \min_{\hat{\bm{b}}_{t}}   ~&  T_t^{\rm{F}}(\hat{\bm{b}}_{t}|\bm{x}_{t},b_{0,t})\nonumber \\
    \text{s.t. } &~\text{C3, C5}.
\end{aligned}
\end{equation} 
According to \eqref{eq:FL_delay} and \eqref{eq:SL_delay}, the objective functions of (P6) and (P7) are derived as follows:
\begin{equation}
    \begin{aligned}
        T_t^{\rm{S}}(\bm{l}_{t}|\bm{x}_{t},b_{0,t}) & = \sum_{k\in\mathcal{K}_t^{\rm{S}}} T_{k,t}^{\rm{S}}({l}_{k,t}|{x}_{k,t},b_{0,t}),\\
        T_t^{\rm{F}}(\hat{\bm{b}}_{t}|\bm{x}_{t},b_{0,t}) & = \max_{k\in\mathcal{K}_t^{\rm{F}}} T_{k,t}^{\rm{F}}({{b}}_{k,t}|{x}_{k,t},b_{0,t}).
    \end{aligned}
\end{equation}

\begin{algorithm}[t]  
	\caption{SL-device Bandwidth Allocation}
	\label{alg:SLBand} 
	\begin{algorithmic}[1]
		\STATE Initialize $b_{\rm{L}}=0$ and $b_{\rm{U}}=1$; 
		\REPEAT
            \STATE $b_{0,t}^*=(b_{\rm{L}}+b_{\rm{U}})/2$;
    		\STATE Compute $T_t^{\mathrm{S}}(\bm{l}_{t}^*|\bm{x}_{t},b_{0,t}^*)$, where $\bm{l}_{t}^*$ is obtained from \eqref{eq:spliting_SL};
            \STATE Compute $T_t^{\mathrm{F}}(\hat{\bm{b}}_{t}^*|\bm{x}_{t},b_{0,t}^*)$, where $\hat{\bm{b}}_{t}^*$ is obtained from Algorithm \ref{alg:FLBand};
            \IF{$T_t^{\mathrm{S}}(\bm{l}_{t}^*|\bm{x}_{t},b_{0,t}^*)>T_t^{\mathrm{F}}(\hat{\bm{b}}_{t}^*|\bm{x}_{t},b_{0,t}^*)$}
                \STATE $b_{\rm{L}} = b_{0,t}^*$;
            \ELSE
                \STATE $b_{\rm{U}} = b_{0,t}^*$;
            \ENDIF
        \UNTIL{$| T_t^{\mathrm{S}}(\bm{l}_{t}^*|\bm{x}_{t},b_{0,t}^*)-T_t^{\mathrm{F}}(\hat{\bm{b}}_{t}^*|\bm{x}_{t},b_{0,t}^*) | \leq \epsilon_3$}
        \STATE Output $b_{0,t}^*$ as the optimal solution of (P5).
	\end{algorithmic} 
\end{algorithm}

Building on this, we proceed to solve (P5)-(P7) as detailed below:
\begin{itemize}
    \item SL-device model splitting: It can be seen from (P6) that each SL device $k$ determines its optimal $l_{k,t}^{*}$ by minimizing $T_{k,t}^{\rm{S}}(l_{k,t}|x_{k,t},b_{0,t})$:
    \begin{equation}
        l_{k,t}^{*} = \arg \min_{l_{k,t}\in\mathcal{L}} T_{k,t}^{\rm{S}}(l_{k,t}|x_{k,t},b_{0,t}),\forall k\in\mathcal{K}_t^{\rm{S}}.
        \label{eq:spliting_SL}
    \end{equation}
    Since the computational complexity of exhaustive search grows linearly with the number of cut-layers $L$, it is employed in this work as a practical approach with acceptable complexity.
    \item FL-device bandwidth allocation: $T_{k,t}^{\rm{F}}(b_{k,t}|x_{k,t},b_{0,t})$ is monotonically decreasing with respect to $b_{k,t}$. Therefore, by contradiction, it can be shown that the optimal value to (P7) is achieved when all FL devices experience the same FL delay, as follows:
    \begin{equation}
    \begin{aligned}
        d^* \!\!& = T_{k,t}^{\rm{F}}(b_{k,t}^*|x_{k,t},b_{0,t})\\
        \!\!& = d_{k,t}^{\rm{FMD}} +d_{k,t}^{\rm{FMT}} + \frac{S}{b_{k,t}^*B \text{log}_2(1+\frac{p_{k} h^{\rm{U}}_{k,t}}{\sigma b^*_{k,t}B})},\forall k\in\mathcal{K}_t^{\rm{F}}.
    \end{aligned}
    \label{eq:bandwidth_FL}
    \end{equation}
    To effectively obtain the optimal $b_{k,t}^{*}$, we propose Algorithm 2, which performs a bisection search to find $d^*$. Notably, $\epsilon_2$ is a small positive constant that controls the precision of the algorithm. 
    \item SL-device bandwidth allocation: $T_t^{\rm{S}}(\bm{l}_{t}^*|\bm{x}_{t},b_{0,t})$ and  $T_t^{\rm{F}}(\hat{\bm{b}}_{t}^*|\bm{x}_{t},b_{0,t})$ in \eqref{eq: delay_P3} are monotonically decreasing and increasing with respect to $b_{0,t}$, respectively. Therefore, the optimal solution to (P5) is achieved when the following condition is satisfied:
    \begin{equation}
        T_t^{\rm{F}}(\hat{\bm{b}}_{t}^*|\bm{x}_{t},b_{0,t}^*) = T_t^{\rm{S}}(\bm{l}_{t}^*|\bm{x}_{t},b_{0,t}^*).
    \end{equation}
    In light of this, we propose Algorithm 3 to efficiently obtain the optimal $b_{0,t}^*$ using bisection search, where a small positive constant $\epsilon_3$ is used to control the precision of the algorithm.
\end{itemize}

\subsubsection{Learning Mode Selection Design}

Through Algorithm 3, (P5), an equivalent reformulation of (P4), is solved with high efficiency. By substituting the obtained optimal value into (P3), we proceed to address (P3) to determine the optimal learning mode for each device. Specifically, we leverage Gibbs sampling to iteratively update the learning mode until convergence to the optimal solution, as summarized in Algorithm 4. In Step 6, a new learning mode is explored with the following probability:
\begin{equation}
\label{eq:exp_prob}
    \epsilon_4 = \frac{1}{1+e^{(\hat{u}_{1t}-u_{1t})/\delta}},
\end{equation}
where $u_{1t}$ and $\hat{u}_{1t}$ are the objective values of (P3) under the current and new learning modes, respectively. The parameter $\delta>0$ controls the exploration tendency. In detail, when $\hat{u}_{1t}-{u}_{1t}<0$, a smaller $\delta$ increases the exploration probability; conversely, when $\hat{u}_{1t}-{u}_{1t}>0$, a smaller $\delta$ decreases it. This behavior encourages updates toward learning modes that yield smaller objective values in (P3), thereby promoting convergence to the optimal solution \cite{Gibbs_Sampling}.


\begin{algorithm} [t]
	\caption{Gibbs Sampling based Learning Mode Selection}
	\label{alg:Gibbs} 
	\begin{algorithmic}[1]
		\STATE Initialize $\bm{x}_t^{\dagger}=\{x_{k,t}| k\in\mathcal{K}\}$ and set $\bm{x}_t^{\mathsection}=\{1-x_{k,t}|k\in\mathcal{K}\}$, where $x_{k,t}$ are randomly generated in $\{0,1\}$; 
        \STATE With $\bm{x}^{\dagger}_t$, 
            obtain objective value of (P3) as $u_{1t}$ using Algorithm \ref{alg:SLBand};
        \REPEAT
    		\STATE Randomly choose one element from $\bm{x}_t^{\dagger}$ and $\bm{x}_t^{\mathsection}$, and swap them with each other to get $\hat{\bm{x}}_t^{\dagger}$ and $\hat{\bm{x}}_t^{\mathsection}$;
            \STATE With $\hat{\bm{x}}_t^{\dagger}$, 
                obtain objective value of (P3) as $\hat{u}_{1t}$ using Algorithm \ref{alg:SLBand};
            \STATE Update $\{\bm{x}^{\dagger}_t,\bm{x}^{\mathsection}_t,{u_t}\}=\{{\hat{\bm{x}}}^{\dagger}_t,{\hat{\bm{x}}}^{\mathsection}_t,\hat u_{t}\}$ with a probability $\epsilon_4$, which is defined in \eqref{eq:exp_prob};
        \UNTIL{Convergence}
        \STATE Output $\bm{x}_t^* = \bm{x}^{\dagger}_t$ as the optimal solution of (P3). 
	\end{algorithmic} 
\end{algorithm}

\subsection{Batch Size Optimization}
To optimize the batch size, we proceed to solve (P2), where the main challenge stems from the complexity of the objective function. To address this, we introduce an auxiliary variable, defined as follows: 
\begin{equation}
\label{eq: tau_def}
    \tau_t = \max\left\{\max_{k \in \mathcal K^{\mathrm F}_t}T_{k,t}^{\mathrm{F}}, \sum_{k\in \mathcal{K}^{\rm{S}}_t}T_{k,t}^{\mathrm{S}}\right\},
\end{equation}
where we have
\begin{equation}
    \begin{aligned}
        T_{k,t}^{\rm{F}} & = \xi_{k,t} \Gamma_{k,t}^{\rm{F}} + \Lambda_{k,t}^{\rm{F}}, \forall k\in\mathcal{K}_t^{\rm{F}},\\
        T_{k,t}^{\rm{S}} & = \xi_{k,t} \Gamma_{k,t}^{\rm{S}} + \Lambda_{k,t}^{\rm{S}}, \forall k\in\mathcal{K}_t^{\rm{S}},
    \end{aligned}
\end{equation}
with $\Gamma_{k,t}^{\rm{F}} =\frac{ C }{f_{k,t}}$, $ \Lambda_{k,t}^{\rm{F}} = d_{k,t}^{\rm{FMD}} + d_{k,t}^{\rm{FMU}}$, $\Gamma_{k,t}^{\rm{S}} =\frac{o_{l_{k,t}}^{\rm{F}}}{R_{k,t}^\mathrm{SU}}+\frac{o_{l_{k,t}}^{\rm{B}}}{R_{k,t}^\mathrm{SD}}+\overset{l_{k,t}}{\underset{l=1}{\sum}}\frac{ c_l}{f_{k,t}} + \underset{l=l_{k,t}+1}{\overset{L}{\sum}}\frac{ c_l}{f_{0,t}} $, and $ \Lambda_{k,t}^{\rm{S}} = d_{k,t}^{\rm{SMD}} + d_{k,t}^{\rm{SMU}}$ as constants.

With the auxiliary variable $\tau_t$, we reformulate (P2) as 
\begin{equation}
\begin{aligned} 
    {\text{(P8)}} \min_{\bm{\xi}_{t},\tau_t} &~\tau_t+\sum_{k\in \mathcal{K}} \frac{\rho_2}{\xi_{k,t}} + \Gamma_2 \nonumber
    \\
    \text{s.t.} \:\: 
    & ~{\text{C}6}\\
    & ~\text{C8}: T_{k,t}^{\rm{F}} \le \tau_t,\forall k \in \mathcal K_t^{\rm{F}} \\ 
    & ~\text{C9}: \sum_{k\in \mathcal{K}^{\rm{S}}_t}T_{k,t}^{\mathrm{S}} \le \tau_t \\ 
    & ~\text{C10}: \tau_t^{\rm{LB}} \le \tau_t \le \tau_t^{\rm{UB}},
\end{aligned}
\end{equation}
where C10 restricts $\tau_t$ within its lower and upper bounds. Since $\tau_t$ increases monotonically with any $\xi_{k,t}$, these two bounds are derived by setting the batch size of each device $k$ to $\xi_{k,t}=1$ and $\xi_{k,t}=D_k$, respectively, as follows:
\begin{equation}
\begin{aligned}
    \tau_t^{\rm{LB}}\!\! & = \max\left\{ \! \underset{{k \in \mathcal K^{\mathrm F}_t}}{\mathrm{max}} \left\{\Gamma_{k,t}^{\rm{F}} + \Lambda_{k,t}^{\rm{F}}\right\}, {\underset{{k\in\mathcal{K}_t^{\rm{S}}}}{\sum}}\left ( \Gamma_{k,t}^{\rm{S}} + \Lambda_{k,t}^{\rm{S}}\right)\!\right\},\\
    \tau_t^{\rm{UB}}\!\! &= \max\left\{\!
        \underset{k \in \mathcal{K}^{\mathrm{F}}_t}{\max} \left\{ D_{k} \Gamma_{k,t}^{\rm{F}} + \Lambda_{k,t}^{\rm{F}} \right\},\!\! 
        \sum_{k \in \mathcal{K}_t^{\rm{S}}}\!\! \left( D_{k} \Gamma_{k,t}^{\rm{S}} + \Lambda_{k,t}^{\rm{S}} \right) \!\right\}.
\end{aligned}
\end{equation}

It can be verified through basic mathematical analysis that (P8) is a convex optimization problem. By introducing the Lagrange multipliers $\bm{\lambda} = \{\lambda_k| k\in\mathcal{K}_t^{\rm{F}}\}$ for C8 and $\mu$ for C9, we derive the Lagrangian dual problem of (P8) as
\begin{equation}
\begin{aligned}
    \text{(P9)} \max_{\bm{\lambda},\mu} & ~g(\bm{\lambda},\mu) \nonumber\\
    \text{s.t. } &~\lambda_k \ge 0, \forall k \in \mathcal{K}_t^{\rm{F}}\\
    &~\mu \ge 0.
\end{aligned}
\end{equation}
The corresponding Lagrangian dual function is given by
\begin{equation}
\label{eq:langDual}
\begin{aligned}
    g(\bm{\lambda},\mu) = \min_{\bm{\xi}_{t},\tau_t} &~  L(\bm{\xi}_{t},\tau_t|\bm{\lambda},\mu)\\
    \text{s.t. } & ~\text{C6 and C9},
\end{aligned}
\end{equation}
where we have
\begin{equation}
     L(\bm{\xi}_{t},\tau_t|\bm{\lambda},\mu) = L_1(\bm{\xi}_{t},\tau_t|\bm{\lambda},\mu) + L_2(\bm{\lambda},\mu),
\end{equation}
with the following definitions: 
\begin{equation}
\begin{aligned}
\label{variable}
    & L_1(\bm{\xi}_{t},\tau_t|\bm{\lambda},\mu) = \tau_t  - \sum_{k\in \mathcal{K}^{\rm{F}}_t} \lambda_k \tau_t -\mu \tau_t+\sum_{k\in \mathcal{K}} \frac{\rho_2}{\xi_{k,t}}\\
    & \qquad \qquad \qquad \quad + \sum_{k\in \mathcal{K}^{\rm{F}}_t} \lambda_k \xi_{k,t} \Gamma_{k,t}^{\rm{F}}+\mu\sum_{k\in \mathcal{K}^{\rm{S}}_t}\xi_{k,t} \Gamma^{\rm{S}}_{k,t},\\
    & L_2(\bm{\lambda},\mu) = \sum_{k\in \mathcal{K}^{\rm{F}}_t} \lambda_k \Lambda_{k,t}^{\rm{F}}+\mu\sum_{k\in \mathcal{K}^{\rm{S}}_t} \Lambda^{\rm{S}}_{k,t} + \Gamma_2.
\end{aligned}
\end{equation}
With the dual variables $\bm{\lambda}$ and $\mu$ fixed, $L_1(\bm{\xi}_{t},\tau_t|\bm{\lambda},\mu)$ becomes a function of the original variables $\bm{\xi}_{t}$ and $\tau_t$, while $L_2(\bm{\lambda},\mu)$ remains constant. Then, we solve the dual problem (P9) to obtain the optimal solution to the original problem (P8), by successively optimizing $\bm{\xi}_{t}$, $\tau_t$, and $\{\bm{\lambda},\mu\}$.

\subsubsection{Original $\bm{\xi}_{t}$ Optimization}

As observed from \eqref{eq:langDual}, C6 defines the lower and upper bounds for each $\xi_{k,t}$ in $\bm{\xi}_t$. Accordingly, the optimal value of $\xi_{k,t}$ can be determined by comparing its stationary point at which the gradient is zero with these bounds. The gradient of $L(\bm{\xi}_{t},\tau_t|\bm{\lambda},\mu)$ with respect to $\xi_{k,t}$ is analyzed separately for FL and SL devices as follows: 
\begin{equation}
    \label{eq: grad_xi}
    \frac{\partial L(\bm{\xi}_{t},\tau_t|\bm{\lambda},\mu)}{\partial \xi_{k,t}} = \begin{cases}
        -\frac{\rho_2}{\xi_{k,t}^2} + \lambda_{k} \Gamma_{k,t}^{\rm{F}}, & \forall k \in \mathcal{K}_t^{\rm{F}}\\
        -\frac{\rho_2}{\xi_{k,t}^2} + \mu \Gamma_{k,t}^{\rm{S}}, & \forall k \in \mathcal{K}_t^{\rm{S}},
    \end{cases}
\end{equation}
from which the stationary point is obtained as
\begin{equation}
\label{eq: sta_point_xi}
    \xi^0_{k,t}(\bm{\lambda},\mu) = \begin{cases}
        \sqrt{\frac{\rho_2}{\lambda_k \Gamma_{k,t}^{\rm{F}}}}, & \forall k \in \mathcal{K}_t^{\rm{F}}\\
        \sqrt{\frac{\rho_2}{\mu \Gamma_{k,t}^{\rm{S}}}}, & \forall k \in \mathcal{K}_t^{\rm{S}}.
    \end{cases}
\end{equation}

For each $\xi_{k,t}$, there are three possible relationships between its stationary point and the bounds specified in C6:
\begin{itemize}
 \item If $1 \leq \xi^0_{k,t}(\bm{\lambda},\mu)\leq D_k$, the stationary point lies within the feasible region, and the optimal solution is given by $\xi_{k,t}^*(\bm{\lambda},\mu) = \xi_{k,t}^0(\bm{\lambda},\mu)$.
 \item If $\xi^0_{k,t}(\bm{\lambda},\mu) < 1$, $L(\bm{\xi}_{t},\tau_t|\bm{\lambda},\mu)$ increases  with $\xi_{k,t}$ over the feasible region. In this case, the optimal solution is attained at the lower bound: $\xi_{k,t}^{*}(\bm{\lambda},\mu)=1$.
 \item If $\xi^0_{k,t}(\bm{\lambda},\mu) > D_k$, $L(\bm{\xi}_{t},\tau_t|\bm{\lambda},\mu)$ decreases monotonically with $\xi_{k,t}$ over the feasible region. Hence, the optimal solution is attained at the upper bound: $\xi_{k,t}^{*}(\bm{\lambda},\mu)=D_k$.
\end{itemize}
Therefore, the optimal batch size for each device $k$ can be summarized as follows:
\begin{equation}
\label{eq: opt_Xi}
    \xi_{k,t}^{*}(\bm{\lambda},\mu) =  \max\left\{1,\min\left \{\xi^0_{k,t}(\bm{\lambda},\mu),D_k\right\} \right\}, \forall k\in\mathcal{K}.
\end{equation}

\begin{rem}
    $\Gamma_{k,t}^{\rm{F}}$ and $\Gamma_{k,t}^{\rm{S}}$ represent the model training delay per data sample for each FL device $k$ and SL device $k$, respectively. From $\eqref{eq: sta_point_xi}$, it can be seen that smaller single-sample model training delays lead to larger batch sizes. Moreover, for each FL device $k$, $\Gamma_{k,t}^{\rm{F}}$ decreases monotonically with the local computing capability $f_{k,t}$, meaning that FL devices with higher computational power are allocated larger batch sizes. For each SL device $k$, $\Gamma_{k,t}^{\rm{S}}$ decreases not only with the local computing capability, but also with the uplink and downlink transmission rates, $R_{k,t}^{\rm{SU}}$ and $R_{k,t}^{\rm{SD}}$, respectively. As a result, SL devices with stronger computing and communication capabilities are assigned larger batch sizes.
\end{rem}

\subsubsection{Original $\tau_t$ Optimization} 
From \eqref{eq:langDual}, we observe that the optimization of $\tau_t$ exhibits the same structural property as that of $\bm{\xi}_t$. Hence, we first identify the stationary point of $\tau_t$, by setting the following gradient to zero: 
\begin{equation}
\label{eq: grad_tau}
    \frac{\partial L(\bm{\xi}_{t},\tau_t|\bm{\lambda},\mu)}{\partial \tau_t} = 1 -\sum_{k\in \mathcal{K}^{\rm{F}}_t} \lambda_k - \mu,
\end{equation}
and then examine its relationship with the lower and upper bounds specified by C10, as detailed below:  
\begin{itemize}
    \item If $\sum_{k\in \mathcal{K}^{\rm{F}}_t} \lambda_k+\mu = 1$, we calculate the stationary point of $\tau_t$ based on its definition in \eqref{eq: tau_def}, as follows:
    \begin{equation}
        \tau_t^0(\bm{\lambda},\mu) = \max \!\left\{\!
            \begin{array}{l} 
                 \underset{k \in \mathcal{K}^{\mathrm{F}}_t}{\max} \left\{ \xi_{k,t}^*(\bm{\lambda},\mu) \Gamma_{k,t}^{\rm{F}} + \Lambda_{k,t}^{\rm{F}} \right\},\\
                 \sum\limits_{k \in \mathcal{K}_t^{\rm{S}}}\!\! \left( \xi_{k,t}^*(\bm{\lambda},\mu) \Gamma_{k,t}^{\rm{S}} + \Lambda_{k,t}^{\rm{S}} \right)
            \end{array}
             \!\right\}.
    \end{equation}
    Since this point lies within the feasible region defined by C10, it also servers as the optimal solution.
    \item If $\sum_{k\in \mathcal{K}^{\rm{F}}_t} \lambda_k + \mu >1$, $L(\bm{\xi}_{t},\tau_t|\bm{\lambda},\mu)$ decreases with $\tau_t$, indicating that the optimal solution is attained at the upper bound: $\tau_t^*=\tau_t^{\text{UB}}$.
    \item If $\sum_{k\in \mathcal{K}^{\rm{F}}_t} \lambda_k + \mu <1$, $L(\bm{\xi}_{t},\tau_t|\bm{\lambda},\mu)$ increases with $\tau_t$. Consequently, the optimal solution lies at the lower bound: $\tau_t^*=\tau_t^{\text{LB}}$.
\end{itemize}

In summary, the optimal learning delay in round $t$ is calculated as
\begin{equation}
\label{eq: opt_tau}
    \tau_t^{*}(\bm{\lambda},\mu) =  
\begin{cases}   
\tau_t^0(\bm{\lambda},\mu), & \sum_{k\in \mathcal{K}^{\rm{F}}_t} \lambda_k + \mu =1 \\  
\tau_t^{\rm{UB}},  & \sum_{k\in \mathcal{K}^{\rm{F}}_t} \lambda_k + \mu >1 \\  
\tau_t^{\rm{LB}}, & \sum_{k\in \mathcal{K}^{\rm{F}}_t} \lambda_k + \mu <1.
\end{cases} 
\end{equation}

\begin{rem}
    Due to the equivalence between (P8) and (P2), the optimal value of $\tau_t$ must follow the expression given in \eqref{eq: tau_def}. As a result, the optimality condition for the dual variables is given by: 
    \begin{equation}
        \sum_{k\in \mathcal{K}^{\rm{F}}_t} \lambda_k^*+\mu^* = 1.
        \label{eq: Opt_Condition}
    \end{equation}
   This condition implies that, in each communication round, the FL delay and the SL delay should be balanced in order to minimize the learning delay $\tau_t$. Specifically, when $\mu$ decreases, larger batch sizes are assigned to SL devices, which in turn increases SL delay. To satisfy \eqref{eq: Opt_Condition}, $\lambda_k$ corresponding to the FL device with the largest learning delay can be increased. This leads to a smaller batch size for this FL device and thus a reduced FL delay. This trade-off aligns the increasing SL delay with the decreasing FL delay to achieve the minimum $\tau_t$. The same reasoning applies in reverse. 
\end{rem}

\subsubsection{Dual $\bm{\lambda}$ and $\mu$ Optimization} 
We adopt the projected subgradient method to update the dual variables $\lambda_k$ and $\mu$:
\begin{equation}
\begin{aligned}
\label{eq: subgrad}
\lambda^{j+1}_k & =\max \left\{0,\lambda^{j}_k + \alpha^{j}_k \Delta_k^{\rm{F}}\right\},\forall k\in \mathcal{K}^{\rm{F}}_t,\\
\mu^{j+1} & =\max \left\{0, \mu^{j} + \beta^{j} \Delta^{\rm{S}} \right\},
\end{aligned}
\end{equation}
where we have
\begin{equation}
\begin{aligned}
    \Delta_k^{\rm{F}} & = \xi^*_{k,t}(\bm{\lambda}^j,\mu^j) \Gamma_{k,t}^{\rm{F}} + \Lambda_{k,t}^{\rm{F}} -\tau_t^*(\bm{\lambda}^j,\mu^j), \forall k\in \mathcal{K}^{\rm{F}}_t,\\ 
    \Delta^{\rm{S}} & = \sum_{k\in\mathcal{K}_{t}^{\rm{S}}}\left(\xi^*_{k,t}(\bm{\lambda}^j,\mu^j)\Gamma_{k,t}^{\rm{S}} + \Lambda_{k,t}^{\rm{S}}\right) -\tau_t^*(\bm{\lambda}^j,\mu^j).  
\end{aligned}
\end{equation}
The step sizes $\alpha^{j}_k$ and $\beta^{j}$ correspond to the updates of $\lambda_k$ and $\mu$ in the $j$-th iteration, respectively. To ensure the convergence of $\lambda_k$ and $\mu$ to the optimal solution of (P9), the step size must satisfy: $\lim_{j \to +\infty} \alpha^{j}_k = 0$ and $\sum_{j=1}^{+\infty}\alpha^{j}_k = +\infty$. The same conditions apply to $\beta^{j}$ \cite{Subgradient_Methods}.

For completeness, we summarize the overall procedure for solving (P9) in Algorithm 5, whose optimal solution is also the optimal solution of (P2). Note that, the termination condition in Step 7 is derived from \eqref{eq: Opt_Condition}, where small positive constant $\epsilon_4$ controls the solution accuracy. 

\begin{rem}
    The $\lambda_k$ update in \eqref{eq: subgrad} implies that when $\Delta_k^{\rm{F}}$ is negative in the $j$-th iteration (i.e., $\xi^*_{k,t}(\bm{\lambda}^j,\mu^j) \Gamma_{k,t}^{\rm{F}} + \Lambda_{k,t}^{\rm{F}} < \tau_t^*(\bm{\lambda}^j,\mu^j)$), $\lambda_k$ is updated to a smaller value. This leads to an increase in the batch size for the corresponding FL device, which in turn raises the term $\xi^*_{k,t}(\bm{\lambda}^{j+1},\mu^{j+1}) \Gamma_{k,t}^{\rm{F}} + \Lambda_{k,t}^{\rm{F}}$ in the $j+1$-th iteration, bringing it closer to $\tau_t^*(\bm{\lambda}^{j+1},\mu^{j+1})$. Conversely, when $\Delta_k^{\rm{F}}$ is positive, $\lambda_k$ is increased, resulting in a smaller batch size and reduced FL delay, again moving toward $\tau_t^*(\bm{\lambda}^{j+1},\mu^{j+1})$. A similar effect occurs in the update of $\mu$, guiding the SL delay toward $\tau_t^*(\bm{\lambda}^{j+1},\mu^{j+1})$. Therefore, the update rule in \eqref{eq: subgrad} facilitates the satisfaction of the optimality condition in \eqref{eq: Opt_Condition} and ensures the convergence of Algorithm 5.
\end{rem}

\begin{algorithm}[t] 
	\caption{Continuous Batch Size Optimization}
	\label{alg:BatchDecision} 
	\begin{algorithmic}[1]
	   \STATE Initialize $j = 1$, $\bm{\lambda}^j = \{\lambda_k^j \mid k \in \mathcal{K}_t^{\rm{F}}\}$, and $\mu^j$;
        \REPEAT
            \STATE Compute $\xi^*_{k,t}(\bm{\lambda}^j, \mu^j)$ using \eqref{eq: opt_Xi};
            \STATE Compute $\tau_t^{*}(\bm{\lambda}^j, \mu^j)$ using \eqref{eq: opt_tau};
            \STATE Update $\lambda_k^{j+1}$ and $\mu^{j+1}$ using \eqref{eq: subgrad};
            \STATE Update $j=j+1$;
        \UNTIL{$|1 - \sum_{k \in \mathcal{K}_t^{\rm{F}}} \lambda_k^{j-1} - \mu^{j-1} | \leq \epsilon_4$} 
        \STATE Output $\bm{\lambda}^{j-1} = \{\lambda_k^{j-1}| k \in \mathcal{K}_t^{\rm{F}}\}$, $\mu^{j-1}$, $\xi_{k,t}^{*}(\bm{\lambda}^{j-1}, \mu^{j-1})$, and $\tau_t^{*}(\bm{\lambda}^{j-1}, \mu^{j-1})$ as the optimal solution to (P9).
	\end{algorithmic} 
\end{algorithm}
\subsection{Batch Size Rounding Algorithm}

By alternately invoking Algorithms 4 and 5 to solve (P1) and (P2), we obtain a locally optimal solution to (P0) without considering C7. This solution provides a lower bound on the optimal value of (P0), denoted as $u_t^{\rm{LB}}$. To satisfy C7, we apply the proposed Algorithm 6 to round the continuous batch sizes to integers, where $\xi_{k,t}^*$ and $\tau_t^*$, being the locally optimal solution to (P0) without C7, are obtained by solving (P8) in the final alternating optimization step of Algorithm 1. First, we take the floor of each batch size to construct feasible integer solution in Step 1, which yields an upper bound on the optimal value of (P0), denoted as $u_t^{\rm{UB}}$. Based on this, we iteratively increase the batch size of the SL device with the smallest batch size until the SL delay constraint is no longer satisfied, as summarized in Steps 2–5. The rationale for adjusting only the SL devices' batch sizes is that the FL delay is already close to $\tau_t^*$; thus, increasing their batch sizes is likely to violate C8. In contrast, SL delay is cumulative and more sensitive to flooring, allowing room for a limited number of additional samples without breaching C9. 

After executing Algorithm 6, an integer batch size solution is obtained. With this fixed, (P1) is solved again to determine the learning mode, model splitting, and bandwidth allocation solution. In this way, the resulting solution is near-optimal to (P0), as its objective value is close to the tight lower bound $u_t^{\rm{LB}}$. This will be validated through experimental results. 

\begin{algorithm}[t]
    \caption{Batch Size Rounding}
    \label{alg:BatchRound}
    \begin{algorithmic}[1]
        \STATE Set $\xi_{k,t} = \lfloor \xi_{k,t}^{*} \rfloor$, $\forall k \in \mathcal{K}$;
        \WHILE{$\sum_{k \in \mathcal{K}_t^{\rm{S}}} ( \xi_{k,t} \Gamma_{k,t}^{\rm{S}} + \Lambda_{k,t}^{\rm{S}} ) < \tau_t^*$}
            \STATE Select SL device $k^* = \arg\min_{k \in \mathcal{K}_t^{\rm{S}}} \xi_{k,t}$;
            \STATE Update $\xi_{k^*,t} = \xi_{k^*,t} + 1$;
        \ENDWHILE
        \STATE Output $\bm{\xi}_t \!\!= \!\{\xi_{k,t}| k \!\in \!\mathcal{K} \}$ as near-optimal solution to (P0).
    \end{algorithmic}
\end{algorithm}

\section{Experimental Results\label{sec:exp_results}}
In this section, we first describe the experimental setup, followed by extensive results and discussions. 

\subsection{Experimental Setup}
We consider a wireless distributed learning system, with system parameters adhering to the setting in \cite{Kun_JSTSP}. In detail, we assume an access point co-located with a server, serving 30 devices randomly deployed within a 100-meter radius. The path loss between each device $k$ and the server depends on the distance ${\rm{dis}}_k$ (in km) between them, modeled as ${\rm{PL}}_k (\rm{dB}) = 128.1 + 37.6 {\rm{log}}_{10}(10^{-3} {\rm{dis}}_k)$. The small-scale fading is assumed to follow normalized Rayleigh fading. The transmit power of the server and each device $k$ are set to $p_{0}=1$ W and $p_{k}= 0.1$ W, respectively. The available bandwidth is set to $B_0=B=1.4$ MHz, with noise power spectrum density as $\sigma=-174$ dBm. The server's computing capability is set to $100\times 10^{8}$ cycles/s, while the device-side computing capabilities are uniformly distributed over the range $[1,8]\times10^8$ cycles/s. Additionally, the number of FLOPs per cycle is set as 16 for the unit conversion between cycles and FLOPs \cite{Linzheng}.

We adopt the CIFAR-10 dataset to train a convolutional neural network (CNN) model for image classification. The data samples are distributed across devices in a non-IID manner. Specifically, we use a Dirichlet-distributed non-IID data partition scheme with a concentration parameter $\phi$ to control inter-device class distribution variance while maintaining intra-device label diversity. A larger $\phi$ indicates a greater degree of non-IID data distribution. Until otherwise specified, we set $\phi=1$ to simulate a medium-level data heterogeneity across devices. The CNN model consists of $L=6$ layers: an input layer, two convolutional layers with $\rm{input\text{ }channel}\times \rm{output\text{ }channel} \times \rm{kernel\text{ }size}$ of $3\times6\times5$ and $6\times16\times5$, and three fully connected layers with $\rm{input\text{ }size}\times \rm{output\text{ }size}$ of $400\times120$, $120\times84$, and $84\times10$. For each convolutional or fully connected layer, the number of parameters in its activations and gradients is equal to the size of the output features, and these are stored in 32-bit floating-point format. The number of FLOPs during backward propagation is twice that of forward propagation, which can be measured using torchstat analyzer in PyTorch. 

For the algorithm parameters, we configure $\epsilon_1 = 10^{-5}$, $\epsilon_2 = 3*10^{-3}$, $\epsilon_3 = 10^{-3}$, and $\epsilon_4 = 10^{-6}$ in Algorithms \ref{alg:alternating}, \ref{alg:FLBand}, \ref{alg:SLBand}, and \ref{alg:BatchDecision}, respectively. Besides, we set $\delta=7.5*10^{-4}$ in Algorithm 4. The values of $\rho_1$ and $\rho_2$ are selected from the sets $\{3, 4, 5, 6, 7, 8, 9\}$ and $\{50, 200, 500, 2000, 5000, 20000, 50000\}$, respectively. Note that, $\rho_2$ can be expressed as a function of $i\in \{3,4,5,6,7,8,9\}$, as follows:
\begin{equation}
    \rho_2 = 5*10^{(i-1)/2}*[i\bmod 2] + 2*10^{i/2}*[(i-1)\bmod 2],
\end{equation}
where $i\bmod2$ denotes the reminder of $i$ divide by 2. For ease of presentation, we denote $\rho_2^{\prime} = i$ as a substitute for $\rho_2$, and simplify $u_t(\bm{x}_{t},\bm{l}_{t},\bm{b}_{t}, \bm{\xi}_{t})$ from \eqref{eq:obj} as $u_t$ throughout the experimental results.

\begin{figure}[t]
    \centering
    \includegraphics[width=\linewidth]{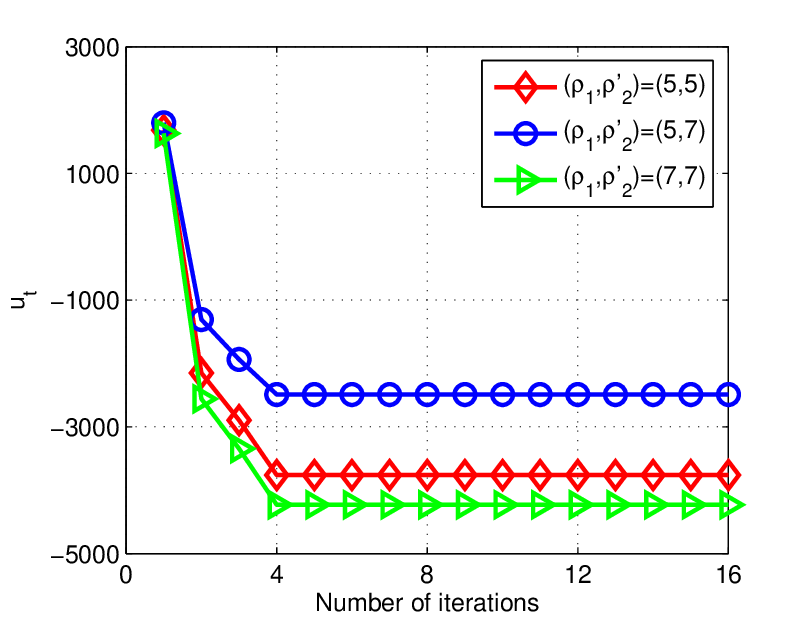} 
    \caption{Convergence of Algorithm \ref{alg:alternating}.}
    \label{fig:coordinate}  
\end{figure}
\begin{figure}[t] 
    \centering
    \includegraphics[width=\linewidth]{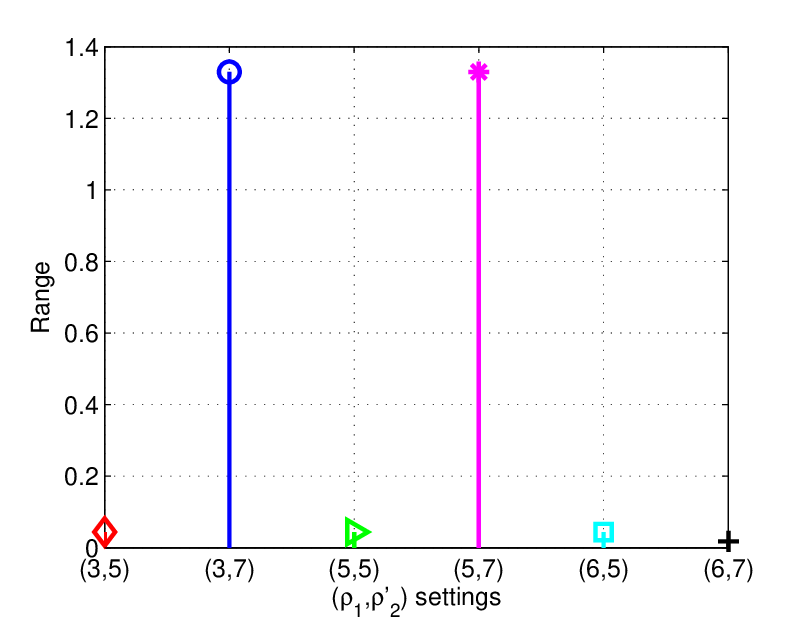}
    \caption{Near-optimality of Algorithm \ref{alg:alternating}.}
    \label{fig:BatchRound}  
\end{figure}

\subsection{Convergence and Near-optimality of Proposed Algorithm}

In Fig. \ref{fig:coordinate}, we illustrate the convergence behavior of Algorithm \ref{alg:alternating}. The results demonstrate that under various ($\rho_1,\rho_2^{\prime}$) settings, Algorithm 1, which employs the block coordinate descent method, consistently converges. This is attributed to the fact that each block subproblem, i.e., (P1) and (P2), is optimally solved using our proposed Algorithms 4 and 5. Moreover, we observe that the converged objective value $u_t$ decreases as $\rho_1$  increases and $\rho_2^{\prime}$ decreases, which aligns with the definition of $u_t$ in \eqref{eq:obj}. To elaborate, we take $(\rho_1,\rho_2^{\prime})=(5,7)$ as a baseline. The optimal solution with this setting remains feasible for (P0) when $\rho_1$ is increased to $7$ while keeping $\rho_2^{\prime}$ fixed at $7$. As a result, the objective value $u_t$ decreases due to the influence of the larger $\rho_1$. Solving (P0) again under the updated $(\rho_1,\rho_2^{\prime})=(7,7)$ setting yields a new optimal solution with a further reduction in $u_t$ compared to the baseline. A similar reduction in $u_t$ is observed when $\rho_2^{\prime}$ is decreased to $5$, i.e., under  $(\rho_1,\rho_2^{\prime})=(5,5)$ setting. 

\begin{figure}[t]
\centering
\includegraphics[width=7.8cm]{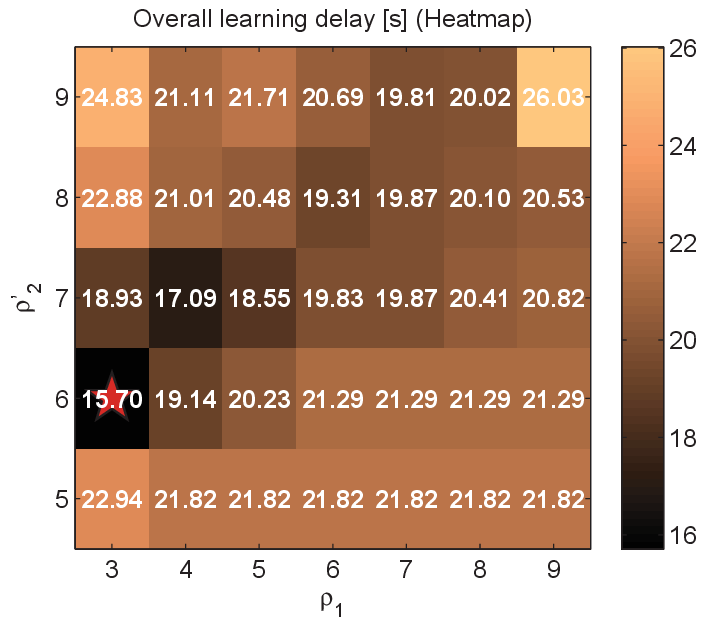}
    \caption{Overall learning delay comparison across different $(\rho_1,\rho_2^{\prime})$ settings.}
    \label{fig:delay_rho1_rho2} 
    \vspace{-0.2cm}
\end{figure}

We demonstrate the near-optimality of Algorithm 1 in Fig. \ref{fig:BatchRound}, where the range, defined as the gap between the upper and lower bounds of the objective value, i.e., $u_t^{\rm{UB}}-u_t^{\rm{LB}}$, is shown. Recall that, $u_t^{\rm{LB}}$ is the optimal value of (P0) when C7 is relaxed, while $u_t^{\rm{UB}}$ is obtained by applying a floor function to the continuous batch sizes. Due to our batch size rounding design in Algorithm 6, the objective value achieved by Algorithm 1 lies within this range. Meanwhile, due to the fact that the true optimal value of (P0) lies within this range as well, the smaller range indicates that the obtained solution is closer to the true optimum. From Fig. \ref{fig:BatchRound}, we observe that the range is extremely small compared to the magnitude of $u_t$ (on the order of thousands, as shown in Fig. \ref{fig:coordinate}). Therefore, the small range under different $(\rho_1,\rho_2^{\prime})$ settings verifies the near-optimality of Algorithm 1.

\subsection{Impact of $(\rho_1,\rho_2)$ Settings on Learning Performance}

\begin{figure}[t]
    \centering
    \subfloat[]{\includegraphics[width=\linewidth]{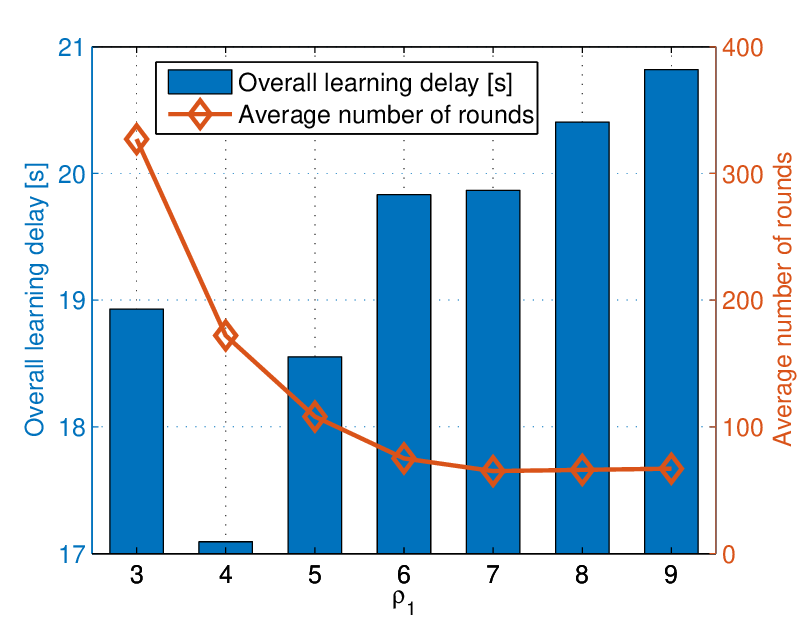}\label{fig:dr_rho1}}

    \subfloat[]{\includegraphics[width=\linewidth]{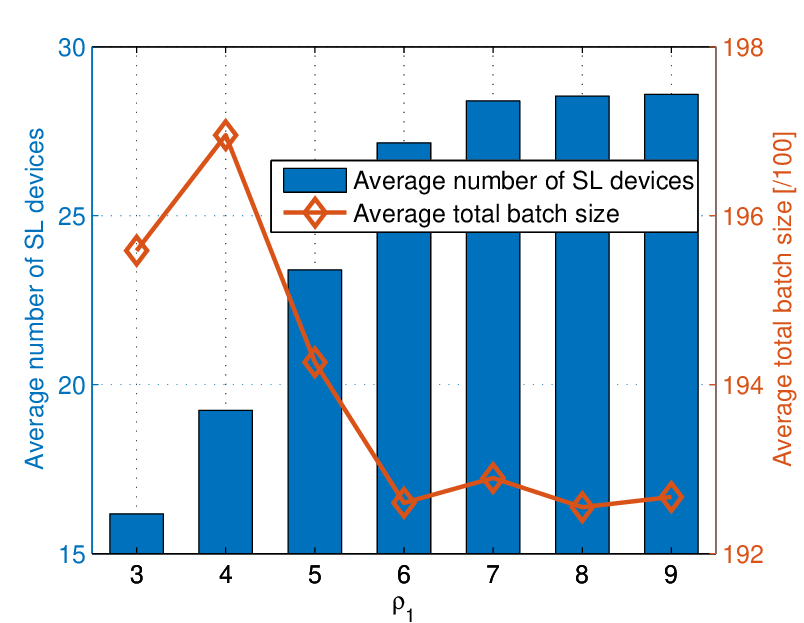}\label{fig:hp_rho1}}
    \caption{Impact of $\rho_1$ on learning performance: (a) Overall learning delay and number of rounds to convergence vs. $\rho_1$; (b) Average number of SL devices and average total batch size per round vs. $\rho_1$.}
    \label{fig:perf_rho1}
\end{figure}

In this subsection, we analyze the impact of the parameter settings $(\rho_1,\rho_2)$, or equivalently, $(\rho_1,\rho_2^{\prime})$, on the overall learning delay, with a target accuracy of 55\%. The experimental results in Fig. \ref{fig:delay_rho1_rho2} clearly show that the interplay between these two parameters plays a crucial role in learning performance. Notably, the overall learning delay exhibits a non-monotonic trend: it first decreases and then increases as both $\rho_1$ and $\rho_2$ increase. This variation in convergence behavior across different $(\rho_1,\rho_2^{\prime})$ combinations underscores the importance of jointly tuning these parameters. Among the configurations tested in Fig. \ref{fig:delay_rho1_rho2}, the lowest learning delay is achieved at $(\rho_1,\rho_2^{\prime}) = (3,6)$. To understand the underlying cause of this effect, we further investigate the interaction between the number of SL devices and the batch sizes in contributing to the observed performance.

Fig. \ref{fig:perf_rho1} shows the impact of $\rho_1$ on learning performance under the setting $\rho_2^{\prime}=7$. As shown in Fig. \ref{fig:dr_rho1}, the number of rounds required for convergence decreases as $\rho_1$ increases, which is consistent with the insight from Theorem 1. However, a reduced number of rounds does not necessarily lead to a shorter overall learning delay. This is because, the one-round delay may increase due to a higher number of SL devices involved in each round, as illustrated in Fig. \ref{fig:hp_rho1}. Consequently, there exists an optimal value of $\rho_1$ that minimizes the overall learning delay by balancing the trade-off between the increasing one-round delay and the decreasing number of rounds. In this case, the minimum occurs at $\rho_1=4$ in Fig. \ref{fig:dr_rho1}. Moreover, Fig. \ref{fig:hp_rho1} shows that as the number of SL devices increases, the total batch size per round tends to decrease, which also aligns with the conclusion of Theorem 1 for a given target accuracy. 

\begin{figure}[t]
    \centering
    \subfloat[]{\includegraphics[width=\linewidth]{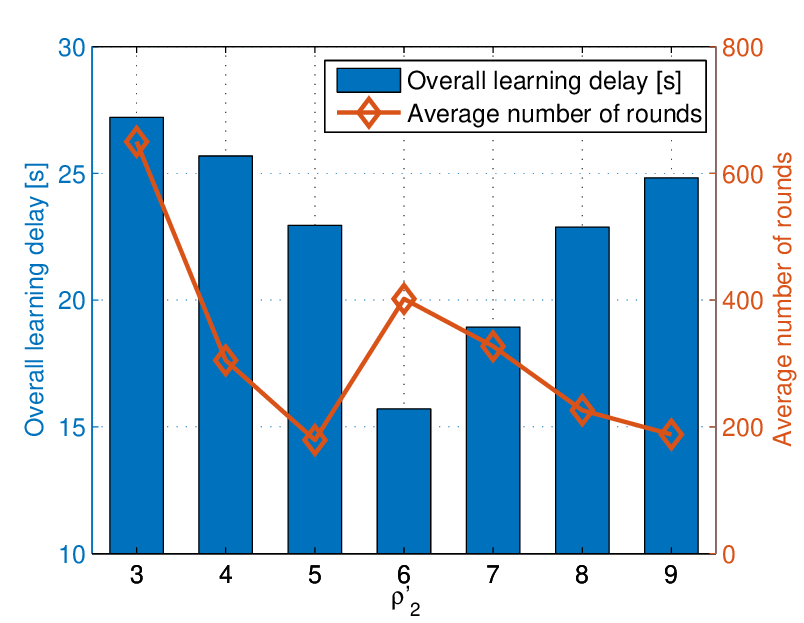}\label{fig:dr_rho2}}

    \subfloat[]{\includegraphics[width=\linewidth]{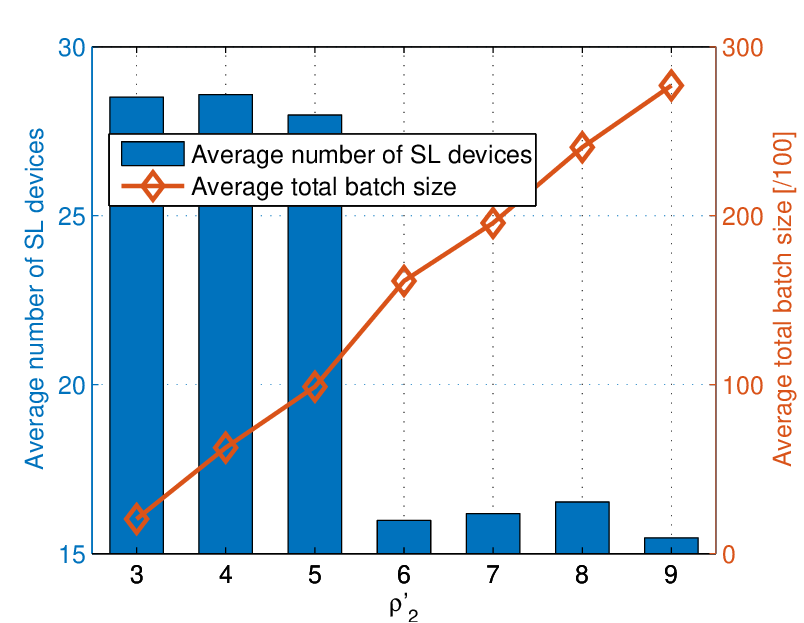}\label{fig:hp_rho2}}
    
    \caption{Impact of $\rho_2^{\prime}$ on learning performance: (a) Overall learning delay and number of rounds to convergence vs. $\rho_2^{\prime}$; (b) Average number of SL devices and average total batch size per round vs. $\rho_2^{\prime}$.}
    \label{fig:perf_rho2}
\end{figure}

With $\rho_1$ fixed at $3$, we examine the impact of $\rho_2^{\prime}$ on learning performance. As $\rho_2^{\prime}$ increases, total batch size per round also increases, which aligns with the calculation in \eqref{eq: sta_point_xi}. This, in turn, leads to a reduction in the number of SL devices, as revealed in Theorem 1. Such an interaction between the total batch size and the number of SL devices per round is illustrated in Fig. \ref{fig:hp_rho2}. In general, a larger total batch size leads to fewer rounds needed to reach the target accuracy. However, as shown in Fig. \ref{fig:dr_rho2}, a noticeable jump occurs at $\rho_2^{\prime}=6$, resulting in a temporary increase in the number of rounds. This anomaly is caused by a sharp drop in the number of SL devices. Interestingly, the decreased one-round delay and increased number of rounds at this jump point together yields the minimum overall learning delay.

\subsection{Learning Performance Comparison with Baselines}
Based on Fig. \ref{fig:delay_rho1_rho2}, we select the optimal setting $(\rho_1,\rho_2^{\prime})=(3,6)$ to showcase the learning performance of HSFL accelerated by the proposed Algorithm \ref{alg:alternating}. For comparison, we consider the following baseline schemes: 

\begin{figure}[t]
    \centering
    \subfloat[]{\includegraphics[width=\linewidth]{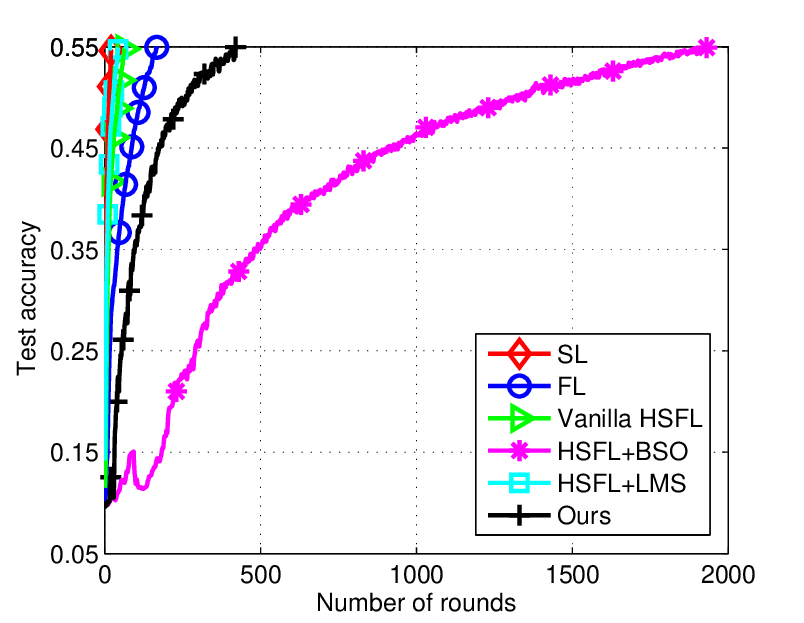}\label{fig:acc_round}}

    \subfloat[]{\includegraphics[width=\linewidth]{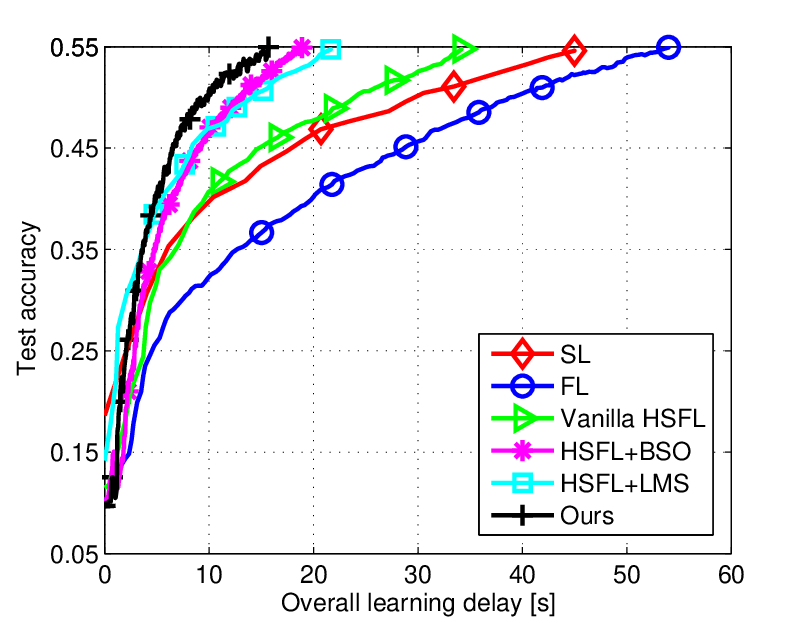}\label{fig:acc_delay}}
    \caption{Learning performance comparison across different schemes: (a) Test accuracy vs. number of rounds; (b) Test accuracy vs. overall learning delay.}
    \label{fig:perf_round_delay}
\end{figure}

\begin{itemize}
    \item SL: In each round, all devices perform split learning sequentially, using a randomly selected cut-layer and the full batch;
    \item FL: In each round, all devices perform federated learning with equal bandwidth allocation and the full batch;
    \item Vanilla HSFL: In each round, all devices randomly participate in HSFL as either FL or SL devices, using the full batch. The total bandwidth is evenly allocated among all devices, and the aggregated bandwidth assigned to SL devices is used sequentially by them, while each SL device randomly selects a cut-layer;
    \item HSFL+BSO: HSFL is enhanced solely by the proposed batch size optimization algorithms (Algorithms 5 and 6), while learning mode selection, model splitting, and bandwidth allocation are the same as in vanilla HSFL; 
    \item HSFL+LMS: HSLF is enhanced using only the proposed learning mode selection algorithm (Algorithm \ref{alg:Gibbs}), without batch size optimization.
\end{itemize}

As shown in Fig. \ref{fig:perf_round_delay}, our proposed algorithm significantly accelerates HSFL by effectively balancing the trade-off between one-round delay and the number of rounds to convergence, ultimately achieving the lowest overall learning delay required to reach the target accuracy. To  further highlight this advantage, Fig. \ref{fig:acc_delay} compares different schemes, revealing that all HSFL variants outperform both FL and SL in terms of learning delay. This is because, in SL, the sequential training across all participants leads to prolonged learning delay. In FL, the need for full model training at all participants exacerbates the delay due to the presence of stragglers with limited computational capacity. In contrast, HSFL leverages a powerful server to attract devices as SL participants, thereby accelerating training in sequential SL. And meanwhile, it reduces the number of FL participants, effectively mitigating the straggler effect. As a result, HSFL achieves a substantially shorter overall learning delay. Furthermore, by jointly optimizing learning modes, model splitting, bandwidth allocation, and batch sizes, the proposed Algorithm \ref{alg:alternating} makes full use of communication and computational resources, enabling rapid convergence to the target accuracy. 

\begin{figure}[t] 
\centering
\includegraphics[width=\linewidth]{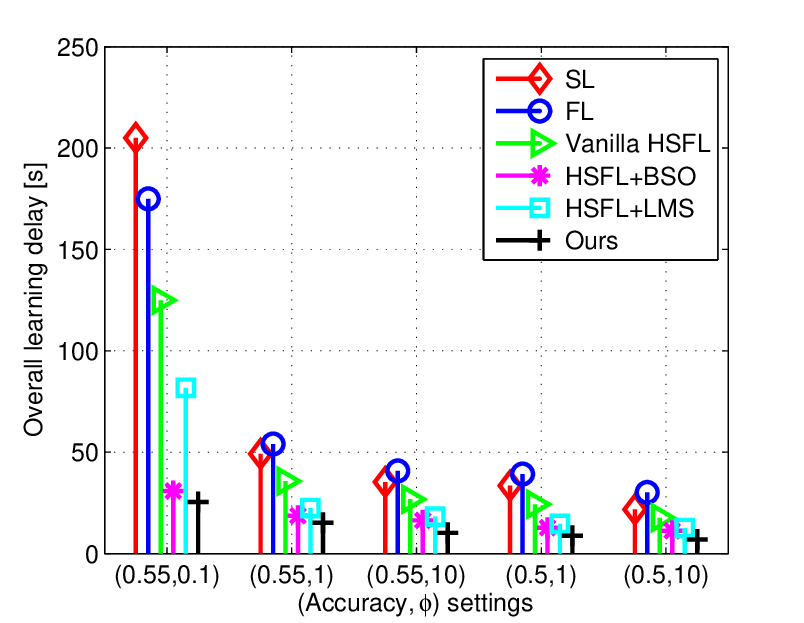} 
\caption{Overall learning delay comparison across different schemes under varying non-IID levels and target accuracy requirements.}
\label{fig:iidCmp}
\end{figure}

Finally, we tune the parameter $\phi$ in the Dirichlet distribution to control the level of data non-IIDness across devices. Specifically, a larger $\phi$ indicates a more severe degree of non-IID data distribution. As shown in Fig. \ref{fig:iidCmp}, our proposed algorithm incurs the lowest overall learning delay across various combinations of target accuracy and $\phi$. Moreover, the overall learning delay decreases as either the target accuracy or $\phi$ is reduced, which aligns with expectations. 

\section{Conclusions\label{sec:conclusions}}
In this paper, we have considered a novel wireless distributed learning framework, termed HSFL, where devices have been allowed to flexibly participate in learning either as FL or SL participants. This design has enabled HSFL to leverage the strengths of both FL and SL to improve learning performance. To further accelerate HSFL under limited communication and computational resources, we have analyzed the impact of learning mode and batch size on convergence, and have jointly optimized these two hyperparameters along with model splitting and bandwidth allocation. This joint optimization has aimed to strike a balance between the one-round delay and the number of rounds to convergence, thereby minimizing the overall learning delay. Extensive experimental results have demonstrated the superiority of the proposed algorithm, especially in terms of reducing the overall learning delay required to reach the target accuracy.

\normalem
\bibliographystyle{IEEEtran}
\bibliography{journal}

\end{document}